\documentclass[journal]{IEEEtran}

\usepackage{float}
\usepackage{gensymb} 
\usepackage{amssymb}
\usepackage{lipsum}
\usepackage{amsmath}
\usepackage{caption}
\usepackage{subcaption}
\usepackage{graphicx}

\begin{document}

\title{IntelliBeeHive: An Automated Honey Bee, Pollen, and Varroa Destructor Monitoring System}

\author{Christian~I.~Narcia-Macias, Joselito~Guardado, Jocell~Rodriguez, Joanne~Rampersad-Ammons, Erik~Enriquez, and~Dong-Chul~Kim
\thanks{C. Narcia-Macias (christian.narcia01@utrgv.edu), J. Guardadoo (joselito.guardado01@utrgv.edu), E. Enriquez (erik.enriquez01@utrgv.edu), and D. Kim (dongchul.kim@utrgv.edu) are with the Department of Computer Science, University of Texas Rio Grande Valley}
\thanks{J. Rodriguez (jocellrod@gmail.com) was with the Department of Biology, University of Texas Rio Grande Valley}
\thanks{J. Rampersad-Ammons (joanne.rampersadammons@utrgv.edu) is with the School of Earth, Environmental and Marine Sciences, University of Texas Rio Grande Valley}}

\maketitle

\begin{abstract}
Utilizing computer vision and the latest technological advancements, in this study, we developed a honey bee monitoring system that aims to enhance our understanding of Colony Collapse Disorder, honey bee behavior, population decline, and overall hive health. The system is positioned at the hive entrance providing real-time data, enabling beekeepers to closely monitor the hive's activity and health through an account-based website. Using machine learning, our monitoring system can accurately track honey bees, monitor pollen-gathering activity, and detect Varroa mites, all without causing any disruption to the honey bees. Moreover, we have ensured that the development of this monitoring system utilizes cost-effective technology, making it accessible to apiaries of various scales, including hobbyists, commercial beekeeping businesses, and researchers. The inference models used to detect honey bees, pollen, and mites are based on the YOLOv7-tiny architecture trained with our own data. The F1-score for honey bee model recognition is 0.95 and the precision and recall value is 0.981. For our pollen and mite object detection model F1-score is 0.95 and the precision and recall value is 0.821 for pollen and 0.996 for ''mite''. The overall performance of our IntelliBeeHive system demonstrates its effectiveness in monitoring the honey bee's activity, achieving an accuracy of 96.28\% in tracking and our pollen model achieved a F1-score of 0.831.
\end{abstract}


\IEEEpeerreviewmaketitle

\section{Introduction}

\label{introduction}
\IEEEPARstart{H}{oney} bees (Apis mellifera) are small insects that play a crucial role in maintaining the balance of ecosystems. They serve as important pollinators, contributing to the pollination of crops worth an estimated 15 billion dollars in the United States alone \cite{usdaHoneyBees}. In today's rapidly advancing technological world, innovative solutions can potentially aid honey bees in overcoming challenges such as parasites and other factors that contribute to the decline of bee colonies. Honey bees are renowned for their role as pollinators, facilitating the reproduction of flowers and fruits through the collection of pollen, which eventually leads to the creation of delicious honey.

Vaorra mites, which are not native to the United States and were introduced from Asia, contribute to the decline of honey bee populations \cite{beekeepVarroa}. Varroa mites survive by feeding on the body fat cells of honey bees and extracting essential nutrients from their bodies \cite{VarroaSamuel}, \cite{parasiticMite2016} as well as transmitting viruses that cause deadly diseases to honey bees \cite{beehealth}. The presence of these ectoparasites can devastate a honey bee colony, and even a colony with minimal signs of infestation has a high likelihood (around 90-95 percent) of collapsing \cite{umnVarroaMites}. This poses significant challenges for beekeepers who invest their time and resources in maintaining honey bee colonies, as a single mite can jeopardize their hives.

Throughout the years of beekeeping, there have been methods developed to control over infestation of varroa mites. Today, many beekeepers have kept traditional methods of checking monthly such as sugar rolls, alcohol washes, or using sticky boards to monitor the bees for mites \cite{PestManagement2022}, \cite{varroa_research}. All of these methods have their pros and cons depending on preference but they are all time-consuming and require manual labor and some approaches are destructive, meaning that the sample used for detecting the infestation levels will not be reintroduced back to the hive \cite{varroa_research}. Therefore, a faster and more effective alternative is essential for monitoring infestation levels for such a time-sensitive issue in order to allow beekeepers to give the proper treatment only when needed to help maintain the bee hive population.

Foraging is another important indicator of the beehives' overall health and is important for beekeepers to monitor. Beekeepers use different methods to monitor the honey bee's foraging activity, one example would be using a pollen trap method that utilizes a mesh screen that has big enough holes for the honey bee to go through but small enough to scrap off pollen from the honey bees’ legs \cite{Pollen_Collection}. This method removes the pollen from the bees' legs for the beekeeper to analyze the amount of pollen that is being brought into the hive from when they forage. Removing pollen from the honey bees' legs is not as efficient as it does not collect enough pollen in the mesh screens, which have an efficiency of 3-43 percent in trapping the incoming pollen, making it ineffective \cite{Pollen_Collection}. This measuring method is inaccurate and removes the nourishment from the honey bees, as they feed on pollen and nectar, which can take a toll on their brood development \cite{Pollen_Collection}, \cite{honeybeePollen2014}.

\section{Related Works}

There are numerous techniques that implement approaches to monitor honey bees' health. \emph{A computer vision system to monitor the infestation level of varroa destructor in a honeybee colony} paper deployed a \emph{Monitoring Unit} with a computer system to record honey bees entering their bee hives using a multi-spectral camera and red, blue, and infrared LED lights to collect footage. They then use computer vision to detect varroa destructors and determine the infestation level of the beehive \cite{BJERGE2019104898}. The objective of this study is to propose an alternative method for assessing the infestation level without harming honey bees, which is commonly done in traditional sampling methods as mentioned previously \cite{PestManagement2022}, \cite{varroa_research}.

\emph{A real-time imaging system for multiple honey bee tracking and activity monitoring purpose is to monitor honey bee behavior} research emphasizes monitoring the activity of honey bees in-and-out activity of the beehive in order to assess honey bee colonies' behavior and the hives overall health when exposed to different concentrations of Imidacloprid pesticides \cite{NGO2019104841}. Their system consists of 2 microcomputers, a Jetson TX2 using background subtraction for object segmentation and honey bee tracking and a Raspberry Pi 3 for environment monitoring using sensors.

The \emph{Automated monitoring and analyses of honey bee pollen foraging behavior using a deep learning-based imaging system} study, aims to provide a better and more efficient alternative to analyze the foraging done by honey bees \cite{NGO2021106239}. This monitoring system also consists of the same two microcomputers but this time for object detection, they used YOLOv3's real-time object detection. Their method proved to be a more effective and reliable tool compared to the conventional pollen trap method previously mentioned.

\emph{Pollen Bearing Honey Bee Detection in Hive Entrance Video Recorded by Remote Embedded System for Pollination Monitoring} developed a non-invasive monitoring system to detect pollen-bearing honey bees. The main focus of this paper was to use their own method to classify pollen-bearing honey bees on an embedded system. Their proposed algorithm wasn't far behind from state-of-the-art classification models but was computationally efficient to be implemented in embedded systems \cite{isprs-annals-III-7-51-2016}. 

The IntelliBeeHive project aims to develop a cost-effective monitoring system using Machine Learning to track honey bees in order to monitor their activity, foraging activity, and varroa mites detection without disturbing the honey bees. This monitoring system is placed at the entrance of the beehive and allows beekeepers to keep track of the beehive's overall activity through an account-based website. For our object detection software, we will be using YOLOv7. YOLOv7 is an object detection model introduced in July 2022 that surpasses all previously known object detection models in speed and accuracy \cite{wang2022yolov7}. YOLOv7 achieved the highest accuracy at 56.8 percent AP at 30FPS or higher depending on the GPU \cite{wang2022yolov7}. 

\section{Hardware}
Our monitoring system is implemented on an NVIDIA Jetson Nano Developer Kit. We chose the NVIDIA Jetson Nano taking several factors into consideration including its affordability (\$99 USD at the time of implementation before the global chip shortage) and performance in computer vision applications compared to other Jetson modules available \cite{nvidia_jetson_nano}\cite{nvidia2021} and the Raspberry Pi. Although the Raspberry Pi is more affordable, it does not have the capability to provide live tracking data. 

\begin{figure}[h!]
  \includegraphics[width=\linewidth]{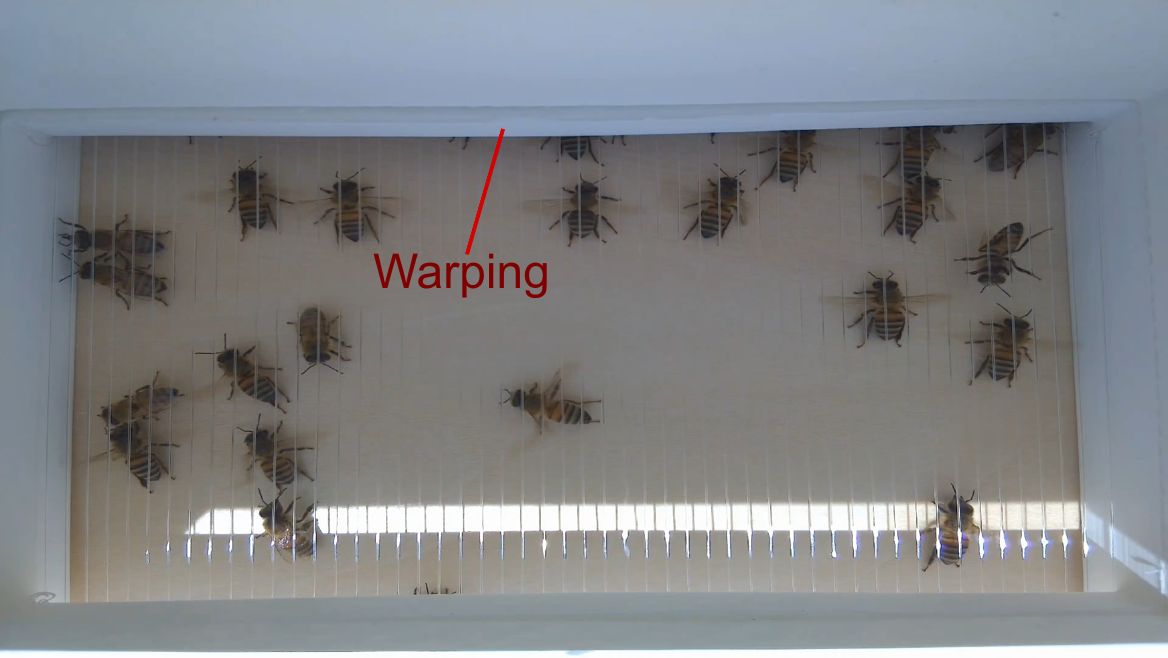}
  \caption[Camera view of fishing line mesh frame warping]{Camera view of fishing line mesh frame warping.}%
  \label{fig:screenMesh}
\end{figure}

The initial design was divided into segments, allowing us to 3D print each section individually. This modular approach facilitated the printing process and provided flexibility to replace specific components if necessary. The container was computer-aid designed (CAD) using Blender, then 3D printed using PLA Filament with three main sections: the Top Box, the Camera Room, and the Mesh Frame. The Top Box has a 3D-printed camera tray to secure a Raspberry Pi Camera, air vents to help cool down the Jetson Nano, and we had to make sure to make it rainproof to protect our electronics, such as the PoE Adapter and the Jetson Nano. The camera room is just an empty box with a window made out of sanded acrylic to reduce glare and allow sunlight to improve inferring accuracy. Our camera distance from the honey bee passage for our PLA container was set at 155 mm high with a viewing area of 150 mm by 80 mm giving us the view shown in Figure \ref{fig:screenMesh}. 

To ensure the effectiveness of our inference algorithm, we devised a method to prevent honey bees from approaching the camera and restricting their movement to prevent overlapping. Our approach involves creating a mesh using a fishing line, as illustrated in Figure \ref{fig:screenMesh}. The use of a fishing line offers several advantages over alternatives such as acrylic. It provides a clearer view of the honey bees without the issue of glare that would occur had we used glass or acrylic. Additionally, using other clear solids would not be viable in the long run, as they would accumulate wax residue and trash over time compromising our tracking algorithm. 

The reason we had to change our 3D printing approach was due to heat and pressure. Over time, we noticed warping with our container in 2 significant locations. One location is where we secured our container to the hive using a bungee cord, the container started to bend inward which in the long run will affect our footage. The second location is the mesh frame, due to the tension caused by the fishing line and the hot temperature in Texas reaching 100 \degree F (37.7 \degree C) during the summer, the mesh frame started to warp inwards loosening the fishing line as shown in Figure \ref{fig:screenMesh} and in return, honey bees are able to break into the camera room compromising our tracking. 

Therefore, we changed our design to laser-cut our container out of wood. While the overall appearance of the container is similar, adjustments in the approach of our CAD design process were made to accommodate the laser-cutting process. In order to laser cut, our 3D model needs to be separated into 2D sections to convert our model into an SVG file format. Using wood gave us a stronger foundation and cut our time to make a container significantly. Previously, the creation of a container took between 4 to 5 days to 3D print, whereas the adoption of laser cutting reduced the time to manufacture to approximately 4 hours followed by an additional day for assembly. The figures below provide an overview of the enclosure and the mesh frame computer-aided design model before converting to SVG.
\begin{figure}[h!]
    \centering
    \begin{subfigure}[b]{0.4\textwidth}
        \includegraphics[width=\linewidth]{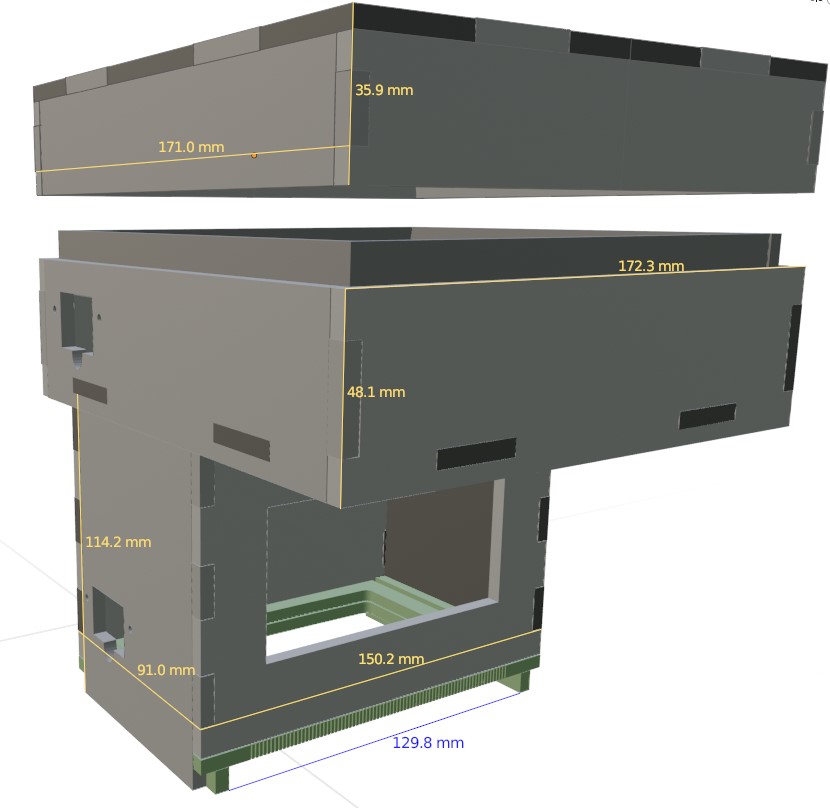}
        \caption[Wooden container CAD assembled overview]{Wooden container CAD assembled overview}
        \label{fig:WoodenContCADOverview}
    \end{subfigure}
    \begin{subfigure}[b]{0.45\textwidth}
        \includegraphics[width=\linewidth]{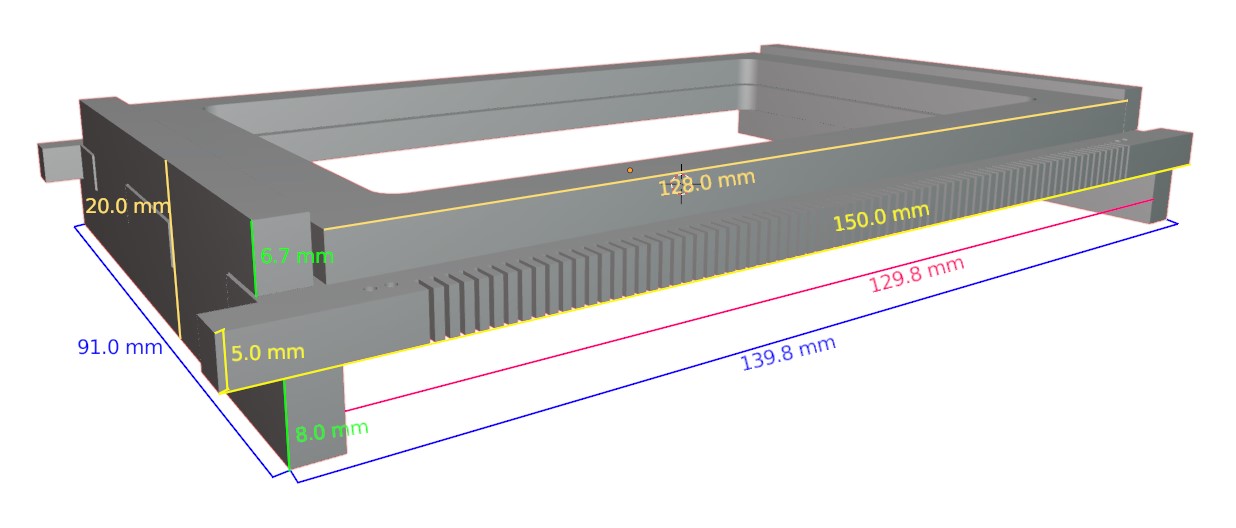}
        \caption[Mesh section CAD Design Overview]{Mesh section CAD design overview}%
        \label{fig:MeshFrameCAD}
    \end{subfigure}
    \caption{CAD enclosure design}
\end{figure}
\begin{figure}
    \centering
    \begin{subfigure}[b]{0.4\textwidth}
        \includegraphics[width=\linewidth]{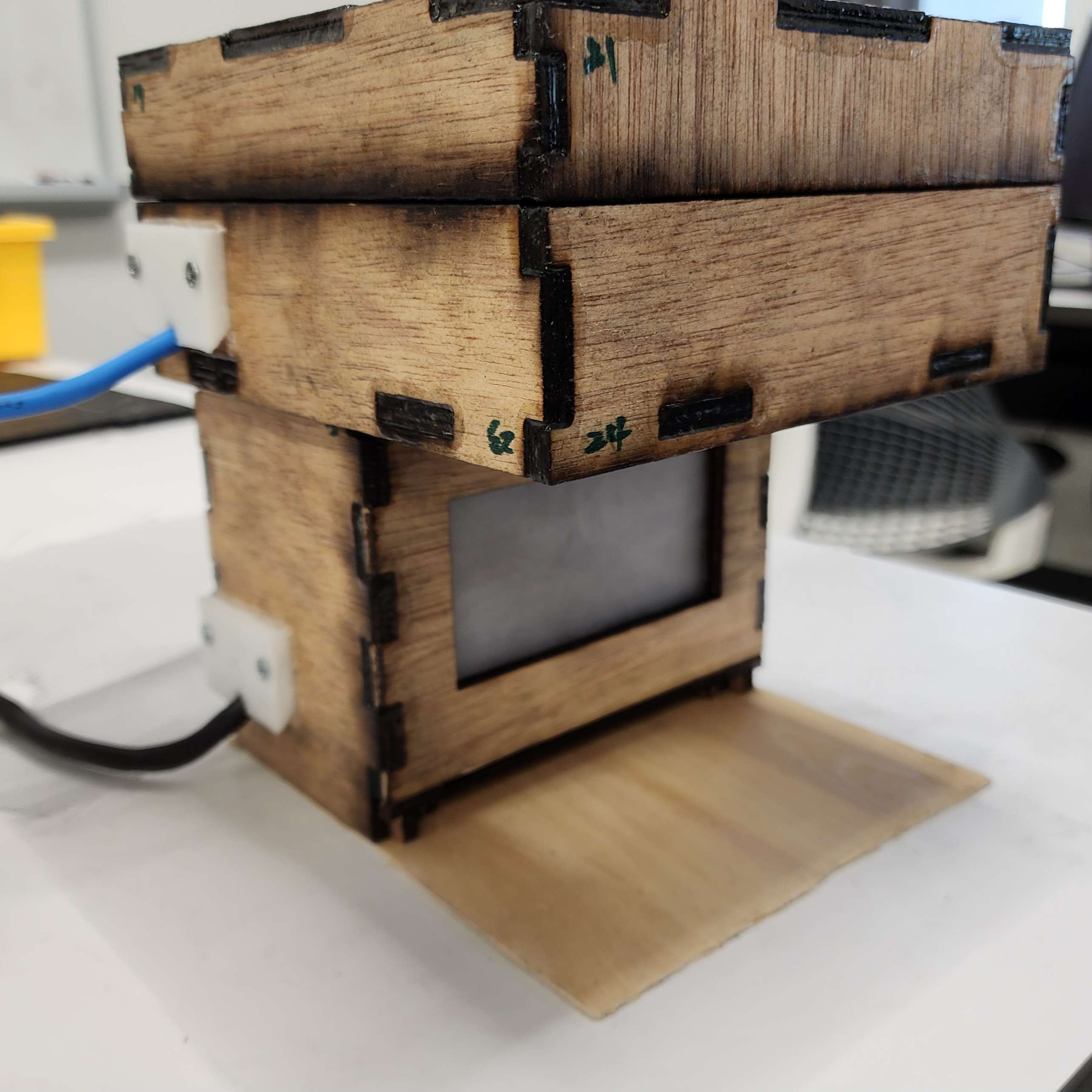}
        \caption[Fully assembled monitoring container]{Fully assembled monitoring container}%
        \label{fig:FullyAssembledCont}
    \end{subfigure}
    \begin{subfigure}[b]{0.4\textwidth}
        \includegraphics[width=\linewidth]{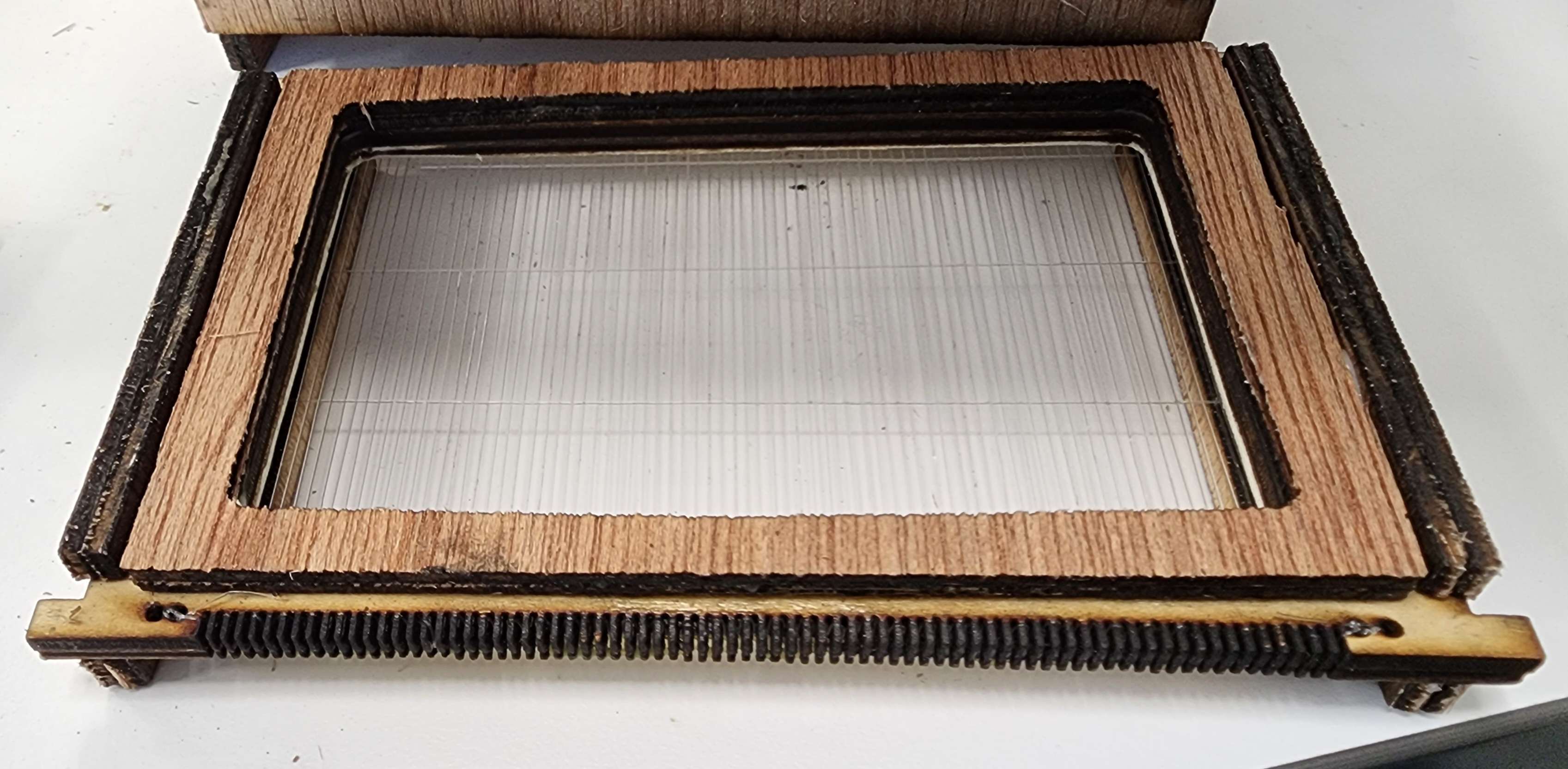}
        \caption[Fully assembled Mesh Frame]{Fully assembled Mesh Frame}%
        \label{fig:FullyAssembledMesh}
    \end{subfigure}
    \caption{Wooden enclosure}
\end{figure}

\begin{figure}[H]
\centering
  \includegraphics[width=.8\linewidth]{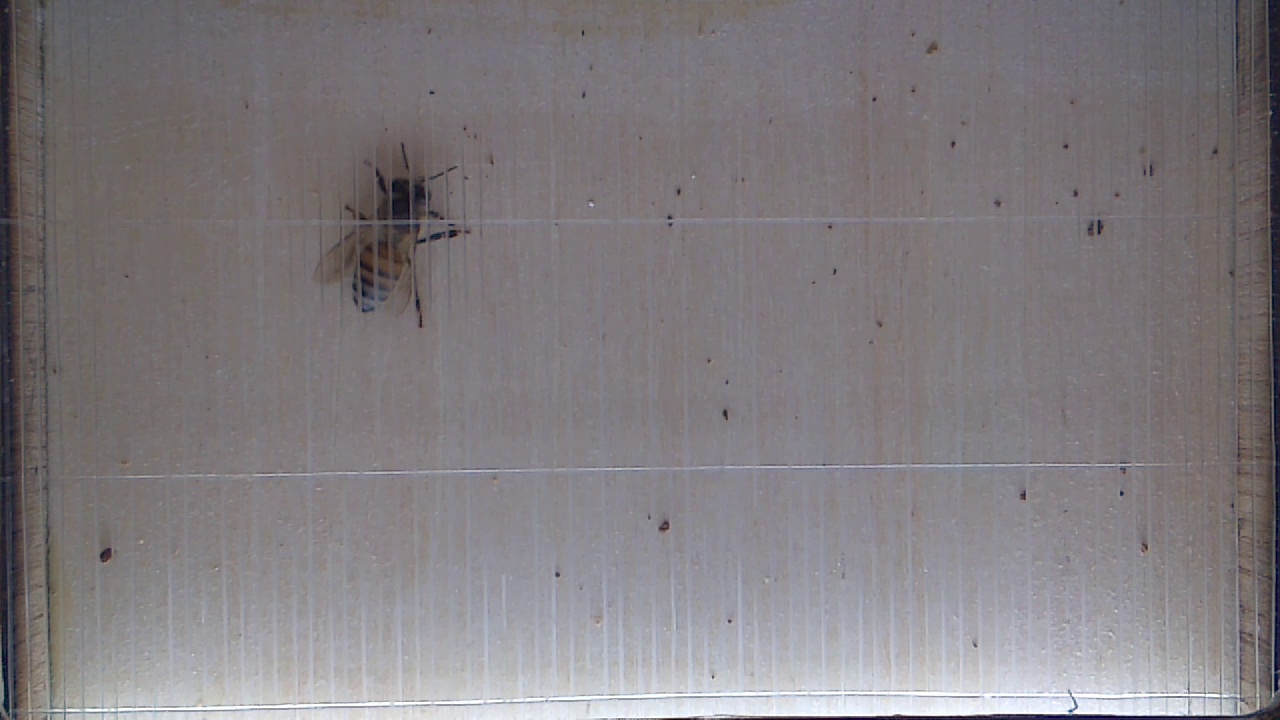}
  \caption[Wooden enclosure camera view]{Wooden enclosure camera view.}%
  \label{fig:WoodenContCameraView}
\end{figure}

Our viewing area for the wooden container was also reduced to allow our camera to get closer to the honey bees improving our pollen and mite detection accuracy. Our new viewing area is reduced to 110 mm by 65 mm and our camera height is lowered to 120 mm giving us a significantly better view of the honey bees as shown in Figure \ref{fig:WoodenContCameraView}.

\begin{figure}[H]
    \includegraphics[width=\linewidth]{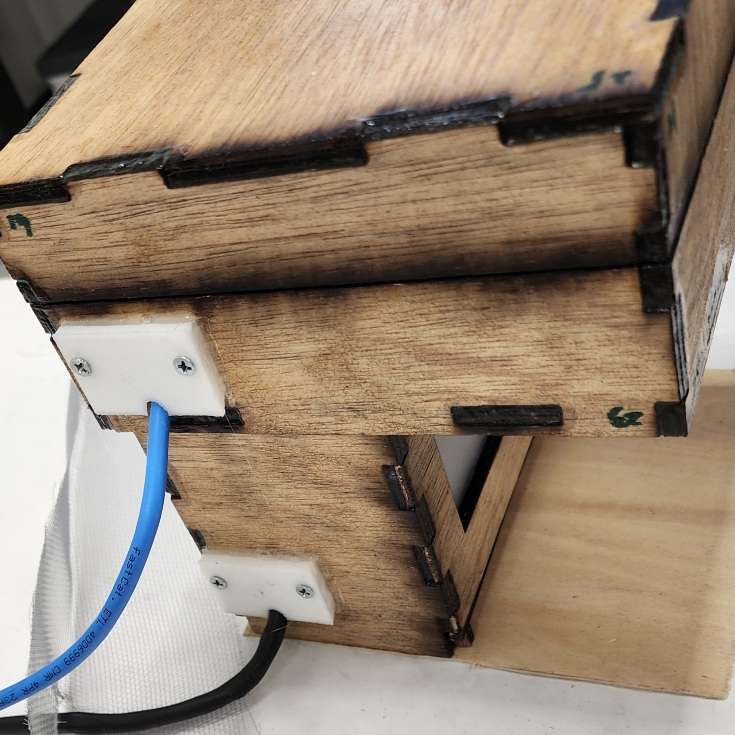}
    \caption[Side View of the container with cable lids attached]{Side View of the container with cable lids attached.}%
    \label{fig:cableCover}
\end{figure}

Our container incorporates two cable exits. The upper cable exit is specifically designated for our Power over Ethernet (PoE) cable, which both powers the Jetson Nano and provides Internet connectivity. The lower cable exit is dedicated to the BME680 sensor, which runs from the top section through the camera room and out into the honey bee hive. In order to achieve a water-tight seal and protect our electronics we use the cable lids we designed shown in Figure \ref{fig:cableCover}.

\begin{figure}[H]
    \centering
    \includegraphics[width=.8\linewidth]{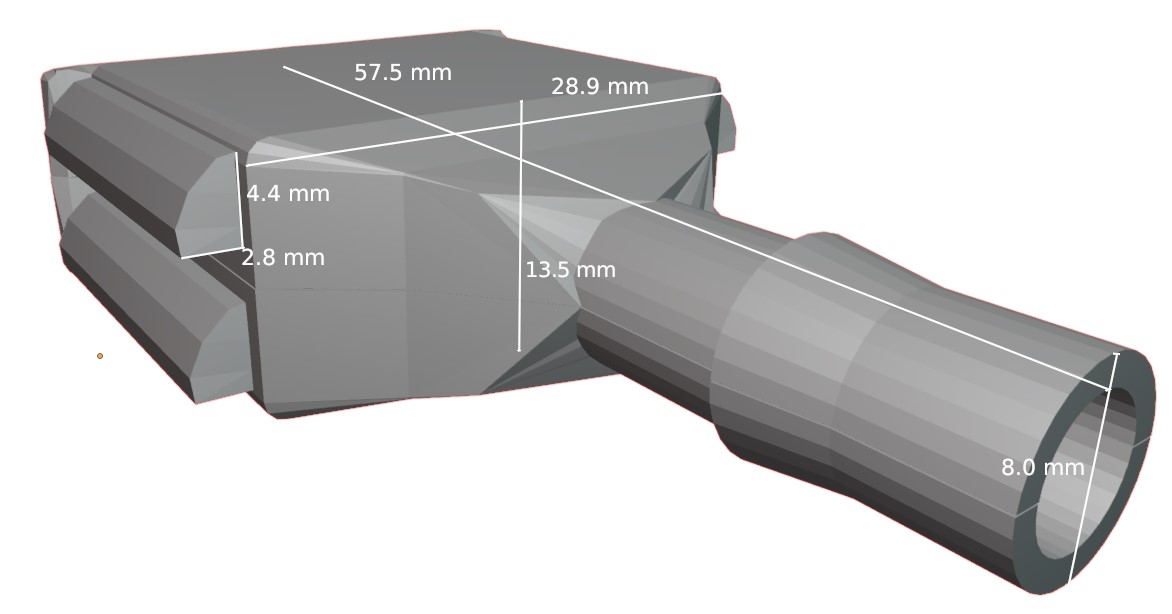}
    \caption[BME680 Sensor Case]{BME680 Sensor Case}%
    \label{fig:SensorCase}
\end{figure}

For monitoring the honey bee hive’s humidity and temperature, we employ a BME680 sensor. Considering this sensor is not specifically intended for outdoor environments, we designed and developed a case with air vents to ensure we don't compromise our readings as shown in Figure \ref{fig:SensorCase}. To 3D print the container we used PLA filament due to its non-toxic nature. To connect our sensor to the Jetson Nano we soldered flexible silicone 30 gauge copper wires to the sensor and ran them through our container to the Jetson Nano’s 40-pin expansion header. We placed the sensor halfway inside the bee hive through the entrance of the bee hive.


\begin{figure}[H]
    \centering
    \includegraphics[width=.4\linewidth]{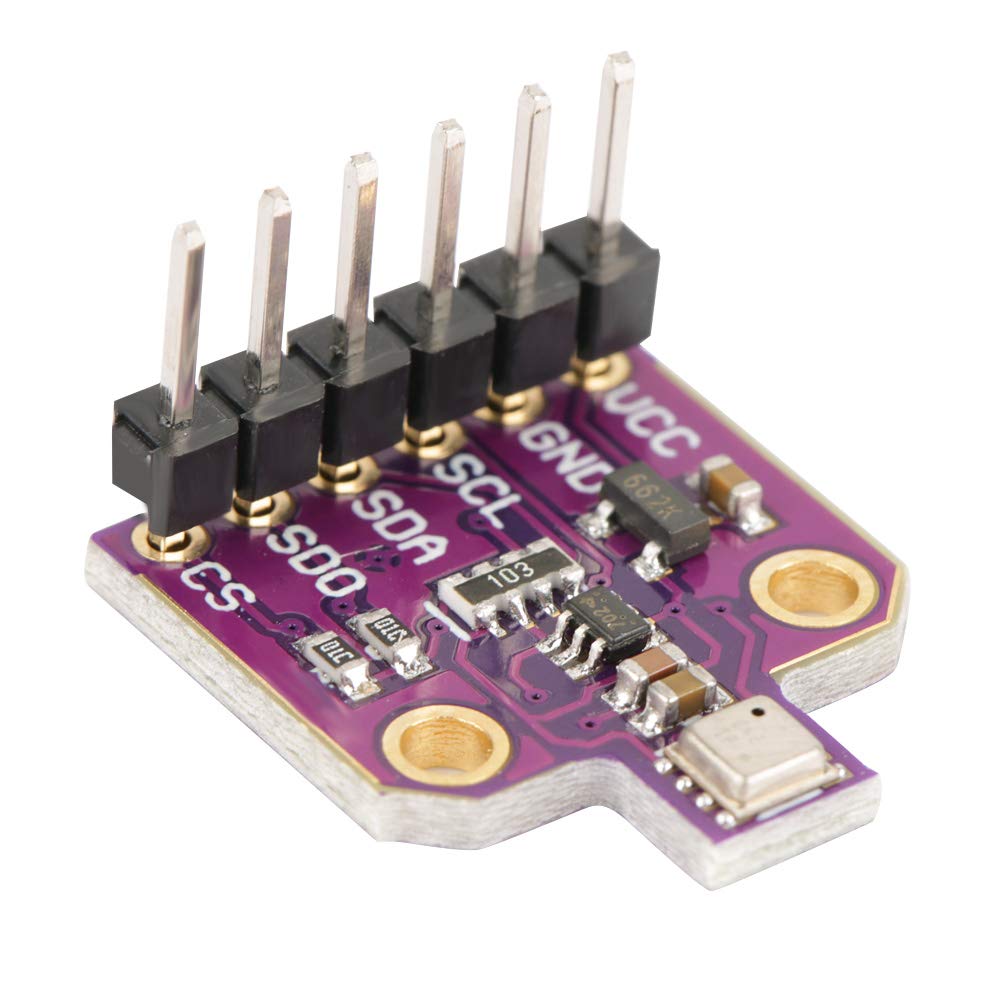}
    \caption[BME680 Sensor]{BME680 Sensor}%
    \label{fig:Sensor}
\end{figure}

\begin{figure}[H]
  \includegraphics[width=\linewidth]{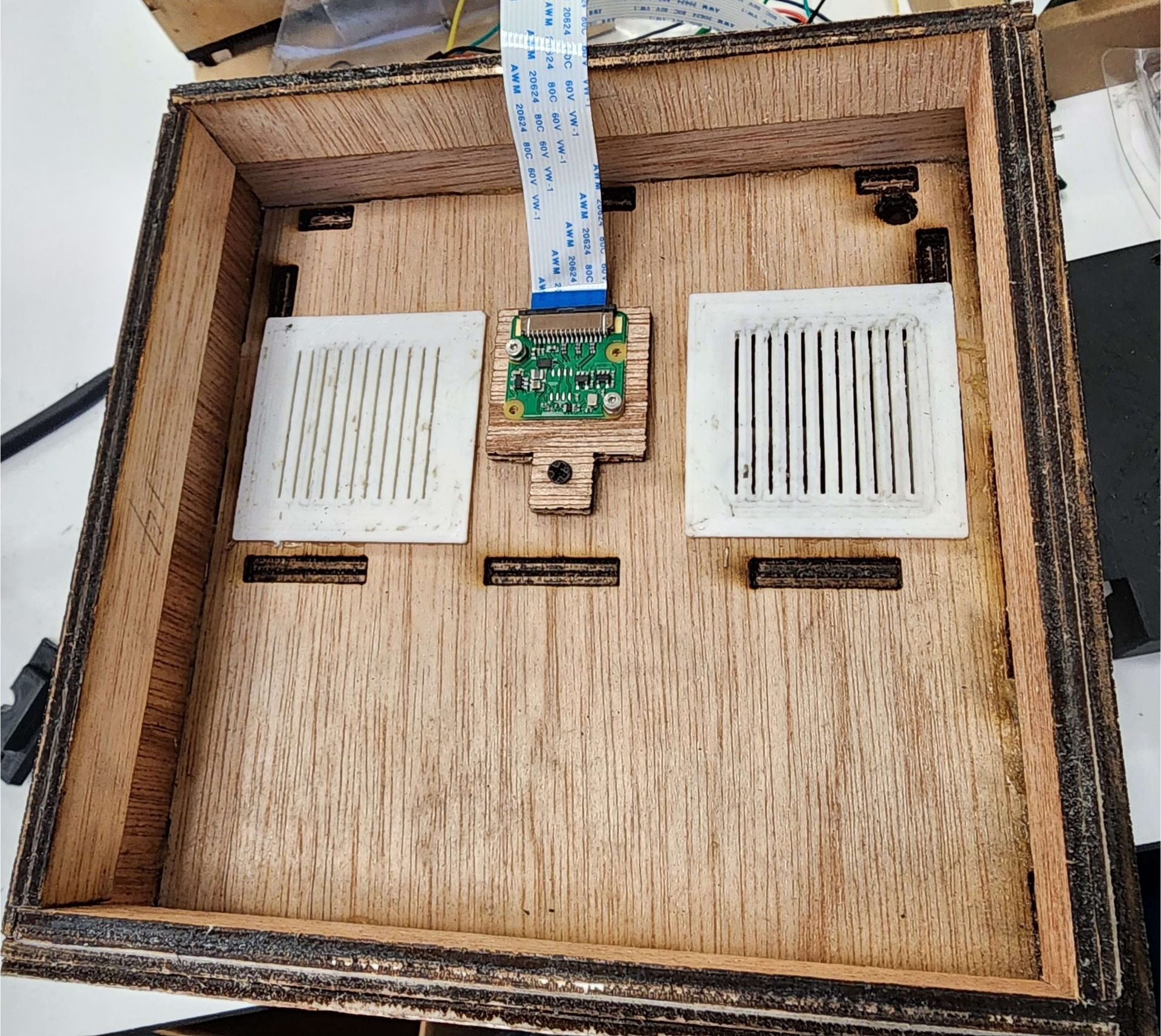}
    \caption[Raspberry Pi Camera V2.1 in monitoring system]{Raspberry Pi Camera V2.1 in monitoring system.}%
    \label{fig:RPICamContainer}
\end{figure}

To capture footage of the honey bee’s in the enclosure, we used the Raspberry Pi Camera V2.1 connected to the Jetson Nano via Raspberry Pi ribbon cable. To hold the camera in place we laser cut a frame from wood and secured it in place in the Top Box as shown in Figure \ref{fig:RPICamContainer}.

\begin{figure}[H]
  \includegraphics[width=\linewidth]{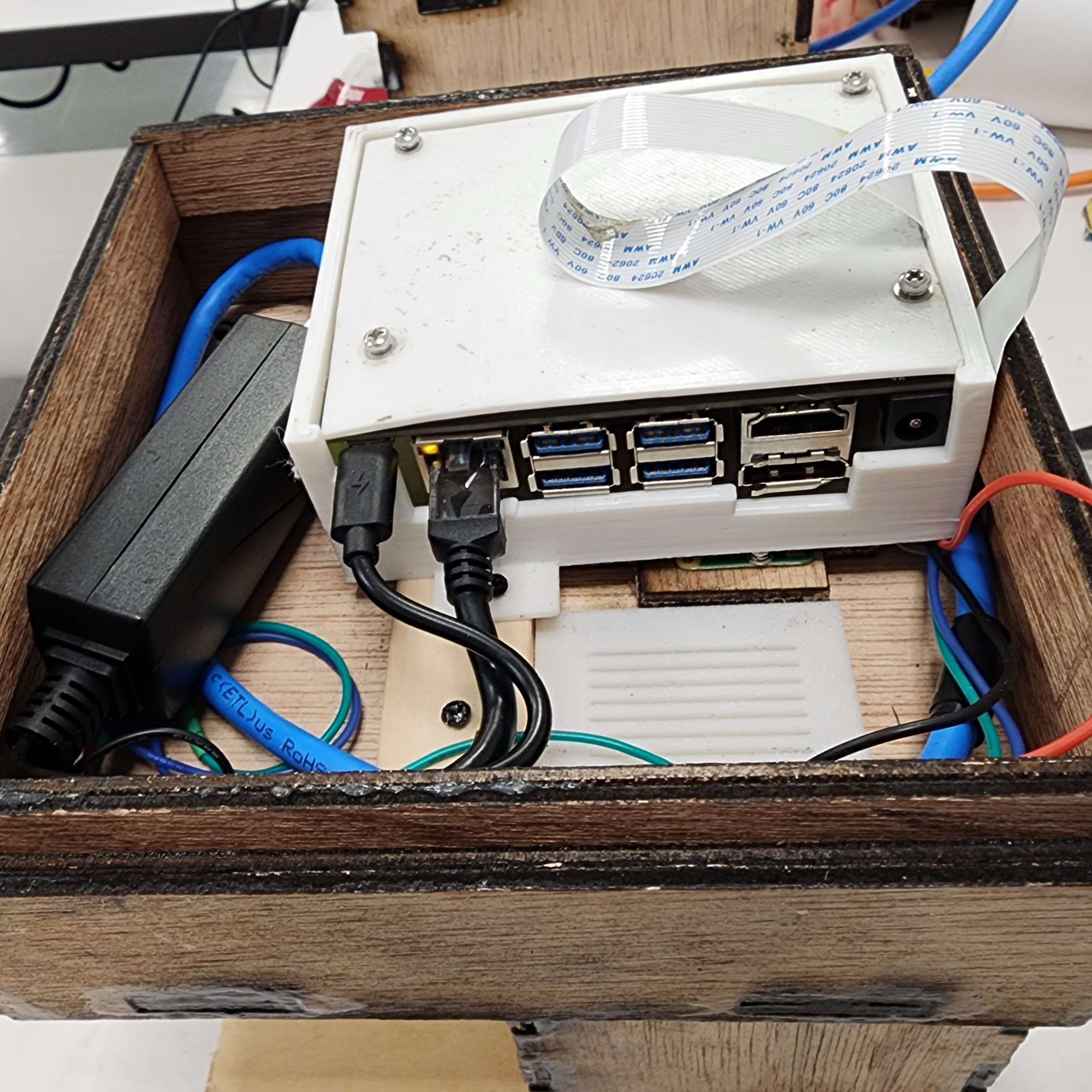}
    \caption[Image of container Top Box section fully assembled]{Image of container Top Box section fully assembled.}%
    \label{fig:TopBoxAssembled}
\end{figure}

To provide internet access and power to our Jetson Nano, we utilize a Power over Ethernet (PoE) switch. A PoE switch provides both power and internet access all through a Cat6 cable running from the PoE switch placed indoors to our PoE adapter inside our container. The PoE Adapter splits the ethernet and power into two channels in order to connect our Jetson Nano. We chose this approach instead of others, such as solar panels, battery packs, or wifi, because it allows us to reduce cable clutter while providing a long-lasting solution with a reliable source of internet and power to our Jetson Nano. Figure \ref{fig:TopBoxAssembled} is an image of the Top Box fully assembled with our Jetson Nano, BME680 sensor cables, Raspberry Pi camera, and PoE adapter all connected.

Lastly, we add a wooden plywood sheet to the bottom of the container. This addition provides a landing place for the honey bees,  gives our object detection a neutral background, and helps stand our container. The wooden plywood can be seen in Figure \ref{fig:FullyAssembledCont}.

\section{Software}

\subsection{Secure Shell Protocol}
In order to enable remote updates for our Jetson Nano device, we implemented Secure Shell Protocol (SSH) tunneling. To ensure accessibility from different networks, we utilized a virtual machines hosted on the Google Cloud platform. This configuration enables us to establish an SSH tunnel from our local computer to the Google Cloud VM, and perform reverse SSH from the Jetson Nano to the Google Cloud.
	
\subsection{Honey bee Detection}

In this study, YOLOv7 Tiny object detection model was used to identify honey bees in order to track their activity. YOLOv7 proved to be the fastest and most accurate real-time object detection model during the implementation of our study\cite{wang2022yolov7}. Due to our computational limitations using a Jetson Nano, we implemented YOLOv7 Tiny version of YOLOv7 to achieve a higher frame rate\cite{wang2022yolov7}.

To train our model, approximately 50 5-minute videos at 10 frames per second at 1280 x 720 every 10 minutes over the span of 4 days (to account for different lighting) were obtained from our own honey bee hive using the containers we developed. Images every 3 seconds (30 frames) were then extracted from the videos to allow the honey bees to move and give us variety in our training data. 

The process of annotating honey bee images for our YOLOv7-Tiny model involved the use of the LabelImg\cite{LabelImg} tool. For our labeling, we purposely annotated only honey bees whose majority of their body is shown in order to improve our detection algorithm due to partial honey bee detection being irrelevant to our tracking and also avoiding flickering if honey bees are on the edge of the frame.  Annotations were saved in the YOLO format with the only class being “Honey bee”, resulting in a total of 1235 annotated images. Approximately 9,700 honey bees were annotated in total. The detection model is trained with an NVIDIA GeForce RTX 3070 GPU. The training image is resized to 416 x 416 pixels input for our YOLOv7-Tiny model with a batch size of 8 for 100 epochs.

Our goal is to have a live status update from every hive with a 5-minute delay. In order to achieve such a goal we must optimize our model as much as possible. 
Given our resource constraints to make our approach cost-effective, our YOLOv7-Tiny model takes approximately 56 ms for every frame for inferring on the Jetson Nano. Since we have a 5-minute video at 10 frames per second totaling 3000 frames, this means that it would take about 2 minutes 48 seconds for inferring only.  
To achieve faster inferring, we convert our model into a TensorRT engine\cite{NvidiaTensorrt}. Before converting our model to TensorRT our model has to be converted into ONNX \cite{ONNX} by exporting our model with the script provided by YOLOv7 repository \cite{YOLOv7}. 

Open Neural Network Exchange (ONNX) is an open standard format that serves as a common representation for machine learning models. It offers a standardized set of operators and a shared file format, allowing AI developers to utilize models seamlessly across various frameworks, tools, runtimes, and compilers. The key benefit of ONNX is its ability to promote interoperability between different frameworks, enabling easier integration and facilitating access to hardware optimizations. By adopting ONNX, developers can leverage the advantages of different frameworks and streamline the deployment of machine learning models\cite{ONNX}. Once our model is in ONNX format, the Tensorrt engine is then created using \emph{TensorRT-For-YOLO-Series} repository\cite{Linaom1214} on the Jetson Nano.
With our TensorRT engine, inferring time was cut by almost half, taking approximately 27 ms per frame. Our total inference time is cut down to about 1 minute and 21 seconds per video.

For our pollen and mite detection, we train a second YOLOv7-Tiny using 2 classes, “Pollen” and “Mite”. To collect pollen training data, we filtered through the videos collected with our container searching for honey bees with pollen. We then extracted the honey bee images for training data from the videos using our YOLOv7-Tiny honey bee detection model. Once we had a collection of approximately 1,000 honey bee images with pollen, we used the Labelimg \cite{LabelImg} tool for annotation.
For mite training data, due to limited time and availability of varroa mites, we used mite placeholders to train our mite detection. We acknowledge that our approach may not perfectly replicate realistic scenarios. However, to simulate the presence of varroa mites on the honey bees, we utilized opaque red beads with a diameter of 1.5 mm as temporary placeholders. While these beads may not accurately mimic the characteristics of actual varroa mites, they served as a substitute to analyze the capabilities of our monitoring system. To collect training data we glued beads onto dead honey bees and extracted data, approximately 700 images of honey bees with "mites". The detection model was also with a NVIDIA GeForce RTX 3070 GPU with the same training parameters except for our input size. For this model our training images were resized to 64 x 64 pixels. Once our YOLOv7-Tiny model was trained, we converted our model into ONNX and then into a Tensorrt engine as we did with our previous model.

\subsection{Tracking Algorithm}

Our tracking algorithm is based on honey bees currently visible. Once the honey bee goes out of sight, it will be counted as a new honey bee if reintroduced. The honey bee’s position is based on the midpoint derived from the detection box extracted from our YOLOv7 tiny model. To track the honey bees we store the current position of each be and compare the previous frame with the current frame to determine if the honey bee moved and in which direction.

Our primary objective is to give as close of a live feed as possible with minimal delay. To achieve this, our monitoring system captures a 5-minute video of the honey bees’ activity and processes the video afterward with our tracking system. While the initial video is being processed, the system concurrently records the subsequent 5-minute video. By adopting this approach, we ensure a near real-time observation of the honey bees' behavior without any significant interruptions.

To record our 5-minute video we use GStreamer recording at 1280 by 720p at 10 frames per second and save our video in 640 by 420p. Downscaling the images is essential to speed up our system's throughput, particularly due to the processing limitations of the Jetson Nano. By downsizing the image, we can significantly enhance the extraction and processing time, resulting in a more efficient workflow. For instance, our processing time for images with a resolution of 1280 by 720p typically takes around 7 minutes and 20 seconds. However, by downscaling, we can reduce this processing time to approximately 3 minutes, excluding the time required for pollen and mite inference. Deepstream can be used to speed up our throughput problem but at the time of implementation, Deepstream isn’t available for Jetpack 4.6 which is the last available Jetpack for Jetson Nanos \cite{nvidia_deepstream}.

Our tracking algorithm uses the output of every frame processed through the honey bee inference TensorRT engine. The output given by our model is based on the upper left and lower right corners of a rectangle of each honey bee inference from the current frame. To determine the midpoint of each honey bee on the video feed we use the following equation:

\[X = (((maxX - minX) / 2) + minX)\]
\[Y = (((maxY - minY) / 2) + minY)\]

The maxX and minY are our coordinates of the lower right vertex of the rectangle and minX and maxY are our upper left vertex. 

To track each honey bee, on initial detection of each honey bee we create a new profile. Each honey bee profile includes Id, last seen location, status, and bee size. To determine whether a honey bee has been detected previously or not when tracking, we use the location of all honey bees detected on frame n-1 and compare them to the output of the current frame n. To consider a honey bee the same bee, we give the new midpoint a tolerance of 50 pixels offset in any direction from the previous location favoring proximity to other honey bees that might be close enough to fall within that range. Any honey bee that does not fall under any currently existing profile is then treated as a new honey bee. Honey bees that don’t have a new midpoint in the current frame are then dropped from the list of active honey bees.

A honey bee can have any of the 4 statuses, “Arriving”, “Leaving”, “New”, and “Deck” depending on their movement. Initially, upon the first detection of the honey bee, they are assigned the status of “New”, meaning that it’s the first time it sees the honey bee or that the honey bee has not crossed any triggers. To track honey bee movement, we have two triggers that change the status of the honey bee. The resolution of the video is set at 640 by 420 pixels meaning the height y of the video is from 0-420 pixels. We then divided the height into three even sections of 140 pixels wide, setting our “Arriving” trigger at 140 pixels, and our “Leaving” trigger at 280 pixels. If the midpoint of the honey bee at n-1 is greater than 140 and n less than or equal to 140, the status of the honey bee changes to “Arriving” meaning that the honey bee is headed to the inside of the beehive, but if the midpoint changes from n-1 is less than or equal to 140 and n greater than 140 the status changes to “Deck” meaning they are in the middle of the container. 

\begin{figure}[H]
    \includegraphics[width = \linewidth]{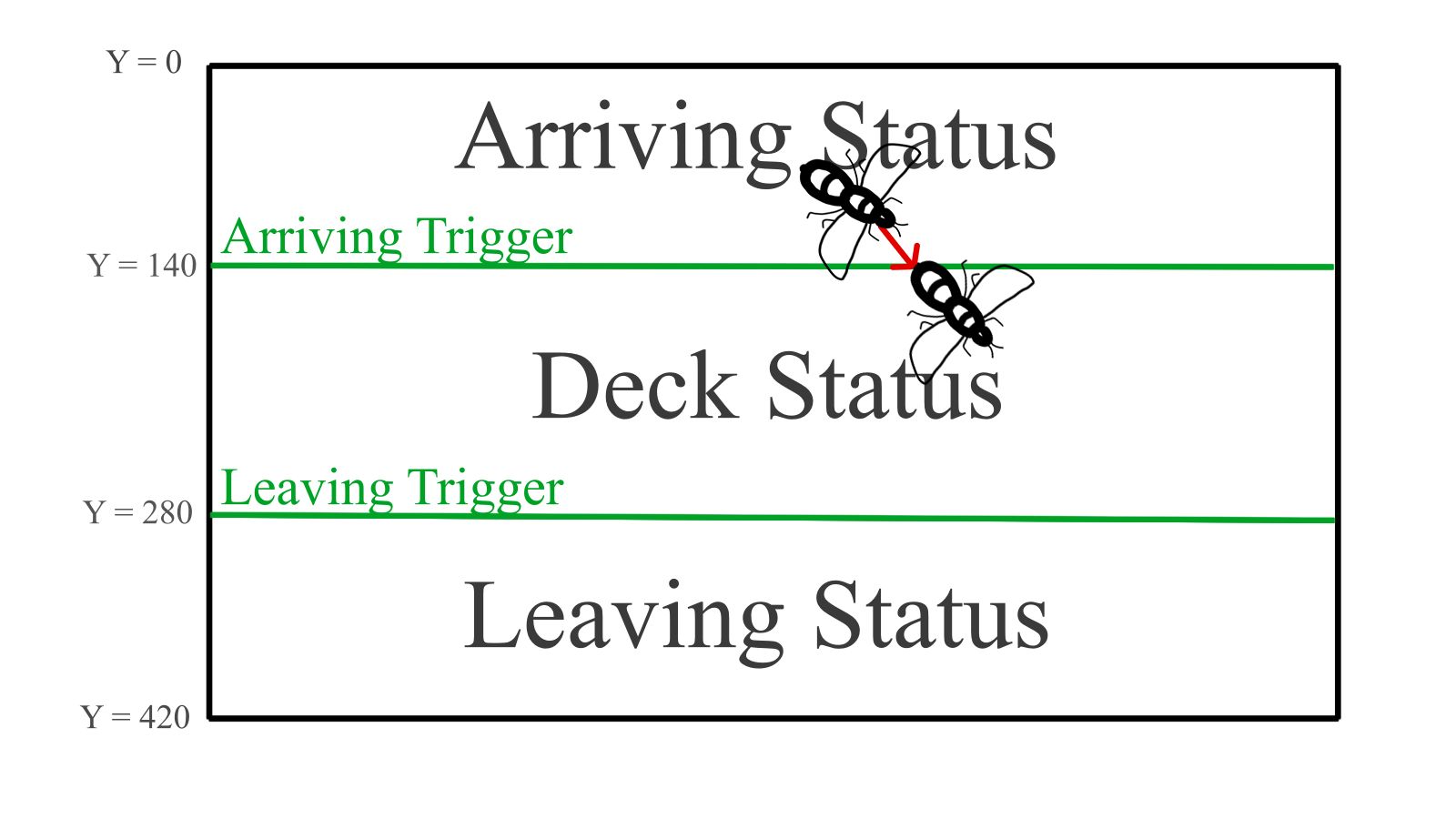}
    \caption[Triggers diagram status breakdown for honey bee tracking]{Triggers diagram status breakdown for honey bee tracking.}%
    \label{fig:TriggerDiagram}
\end{figure}

The "Leaving" trigger is determined based on its crossing at the Y-coordinate value of 280. This trigger will result in the honey bee status being changed to either "Leaving" or "Deck," depending on whether the midpoint is less than 280 at frame n-1 and greater than or equal to 280 at frame n, or if the midpoint is greater than 280 at frame n-1 and less than or equal to 280 at frame n, respectively. Figure \ref{fig:TriggerDiagram} is a diagram demonstrating how the status of the tracking algorithm works.

The honey bee size is extracted once per honey bee profile. The honey bee size is based on the longest side of the rectangle output given by our model. Our camera covers a work area of 110 mm by 65 mm. To get the size of the honey bee we the following formulas:

\[1. (maxX - minX)/(frame size X/ container Size X)\]
\[2. (maxY - minY)/(frame size Y/container Size Y)\]

Formula 1 is used if the longer side of the rectangle is along the X-axis or formula 2 for the Y-axis. We divide the frame size by the container size for the respective axis to get the ratio and determine the size of each bee. The objective of determining the size of each honey bee is to investigate the ratio between a drone and a worker honey bee. However, due to variations in the inference rectangle's size, which can change depending on whether a honey bee is fully visible or not fully present due to it being on the edge of the frame, we only extract the honey bee size when it crosses a "Leaving" or "Arriving" trigger. This approach ensures that we capture the complete size of the honey bee. It is important to note that this method may not be optimal since the size is solely determined by the longest side of the inference rectangle. Consequently, if the honey bee is at an angle when its size is captured, the accuracy and reliability of our data may be affected.

The purpose of considering the honey bee size is to determine if using the size alone is enough to show the difference between worker and drone bees. The graph below shows the size output of our model from a 5-minute video and then manually annotated drone and worker honey bees.

\begin{figure}[H]
    \includegraphics[width = \linewidth]{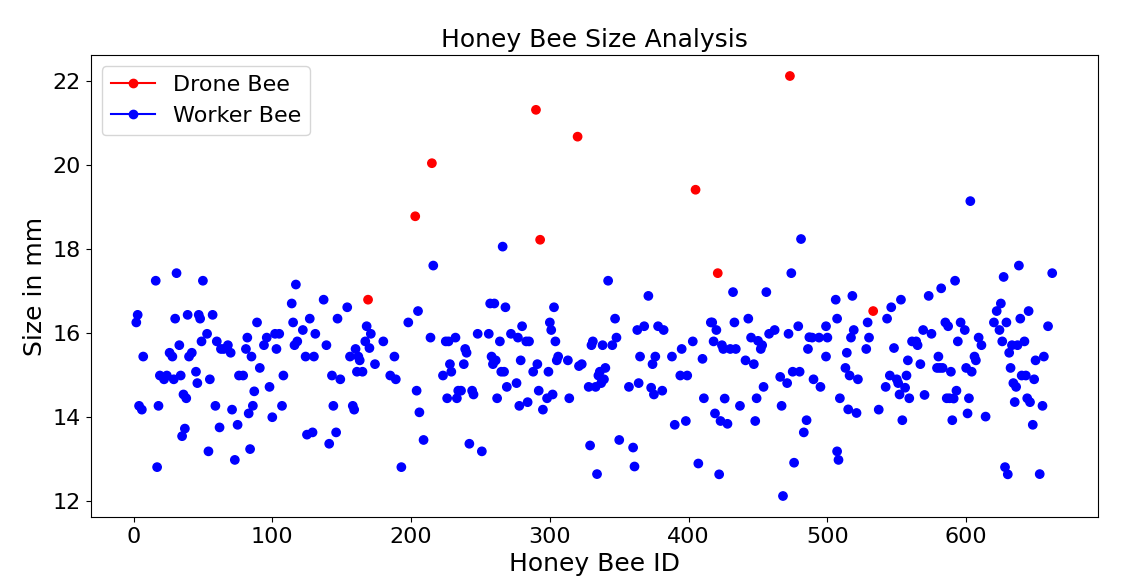}
    \caption[Honey bee drone versus worker bees size analysis]{Honey bee drone versus worker bees size analysis.}%
    \label{fig:Drone_Worker_Analysis}
\end{figure}

The images below are outputs extracted from two profiles of two different types of honey bees inferred from a 5-minute video.

\begin{figure}
    \centering
    \begin{subfigure}[b]{0.22\textwidth}
    \centering
        \includegraphics[scale=1.9]{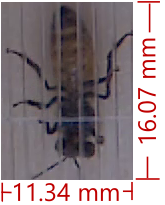}
          \caption[Worker honey bee image extracted from a video frame]{Worker honey bee image extracted from a video frame}%
          \label{fig:HoneyBeeDrone}
    \end{subfigure}
    \hfill
    \begin{subfigure}[b]{0.22\textwidth}
    \centering
       \includegraphics[scale=1.9]{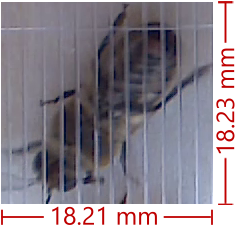}
          \caption[Drone honey bee image extracted from a video frame]{Drone honey bee image extracted from a video frame}%
          \label{fig:HoneyBeeWorker}
    \end{subfigure}
    \caption[Drone and Worker Bee Comparison]{Drone and Worker Bee Comparison}
\end{figure}

To identify the presence of pollen or mites on a honey bee, we follow a specific procedure. For each honey bee profile, we save an image of the honey bee into a designated folder when it passes any of the triggers. This ensures that we capture a complete view of the honey bee for analysis. Once the honey bee TensorRT engine model has completed processing the video, we proceed to load the pollen and mite TensorRT engine model and process all the images extracted by the honey bee TensorRT engine. 

\section{Website}\label{ch:Website}

The IntelliBeeHive has a web application designed to store and present data gathered from honey bee hive monitoring systems, catering to apiarists or beekeepers. Our web page can be found at https://bee.utrgv.edu/. The monitoring system collects hive data, which is then transmitted to the IntelliBeeHive web server via an API. The web server, a remote computer accessible through the internet, receives and stores the data in its database \cite{phpwebapp2000}. An API serves as the interface that enables communication between programs on separate machines \cite{restapi2016}. Once the hive data is stored, it is presented to the user in an organized and user-friendly manner through their web browser whether it'd be on a personal computer or mobile device. This chapter will discuss the functionality of the IntelliBeeHive web application, breaking it down into two main components: the frontend and the backend. The frontend is what the user experiences and interacts with on their personal device, while the backend is what happens on the web server, such as data collection and storage.

\begin{figure}[H]
\centering
  \includegraphics[width=\linewidth]{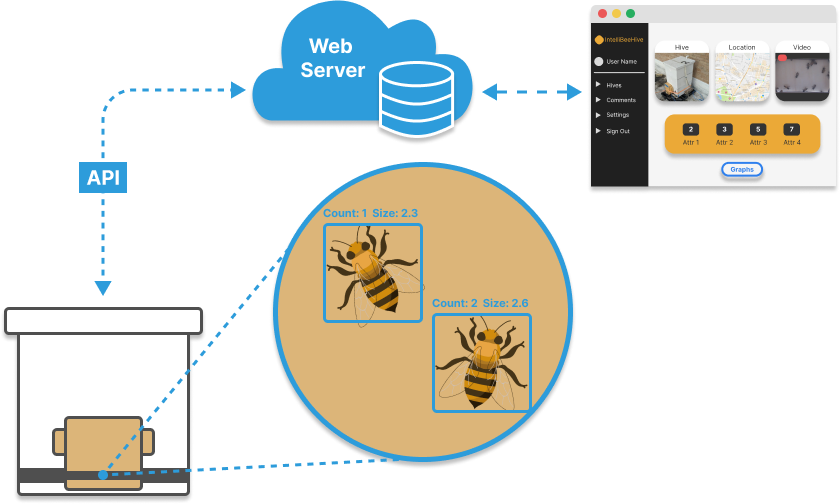}
  \caption[IntelliBeeHive WorkFlow]{Shows an illustration of the IntelliBeeHive web application functionality.}%
  \label{fig:IntelliWorkFlow}
\end{figure}

\subsection{Frontend}
The IntelliBeeHive is designed for apiarists meaning the website is user-friendly and accessible by almost all devices with web access including smartphones and computers. 

\begin{figure}[H]
  \includegraphics[width = \linewidth]{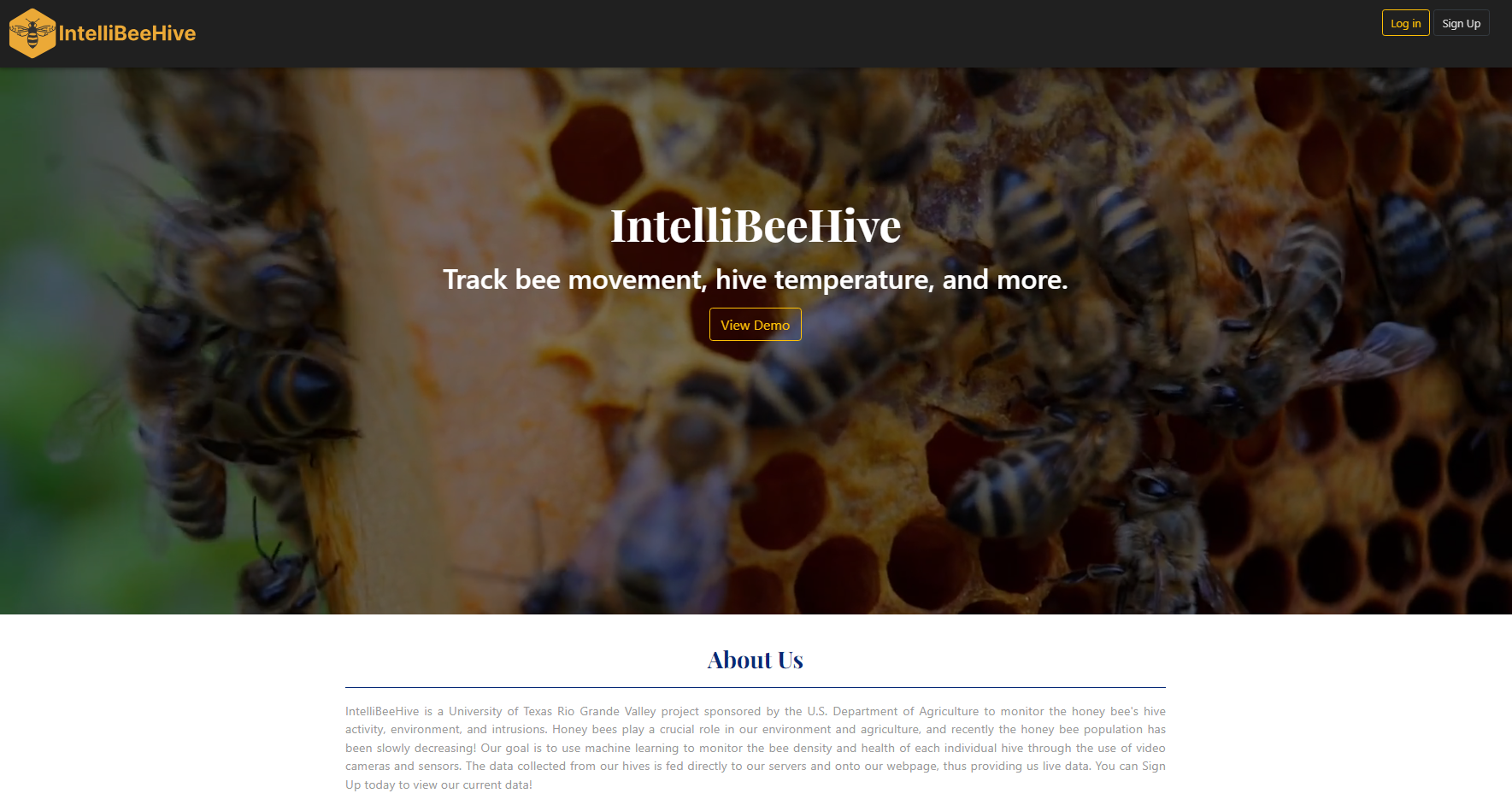}
  \caption[Shows IntelliBeeHive’s landing page welcoming new and current users]{Shows IntelliBeeHive’s landing page welcoming new and current users.}%
  \label{fig:IntelliBeeHiveHome}
\end{figure} 

\subsubsection{Layout}\label{layout}
IntelliBeeHive’s front-end consists of 8 separate web pages. These pages are accessed sequentially and have specific restrictions depending on the type of user accessing them. There are 3 user types: all users, registered users, and admin users.

All users refer to anyone who has access to the IntelliBeeHive website and doesn’t require any credentials. All users have access to the landing, log-in, and sign-up pages and to the hive demo page. The hive demo page displays a single hive's live video recording, data, and statistics.

Once a user signs up and has verified their credentials they become registered users. Registered users have access to the hive feed page, which showcases all hives currently utilizing a monitoring system. The hive feed page provides live and past data in graph and table formats. Registered users can navigate to the comment page to leave feedback or questions regarding the web application. They can also access the settings page to update their credentials or delete their account. 

Registered users can only become admin users if they are granted the privilege by the webmaster. Admin users have special privileges, including the ability to create, edit, and delete hives. They can also view comments submitted by registered users and delete registered user accounts. However, admin users cannot add a monitoring system or link one to an existing hive, as this privilege is exclusive to the webmaster.

\subsubsection{Adding Users}
New users can be added as registered users by signing up through the sign-up page. To complete the sign-up process, users are required to provide their first and last name, email address, and an 8-character alphanumeric password.

The sign-up page will automatically show the user a prompt box where they can input the verification code. For security a user has 24 minutes to input the code before it expires, if the code expires the user will need to start over the sign-up process \cite{phpsession2010}. Once the user inputs the verification code within the specified time limit, their credentials are stored in the web server and they are recognized as a registered user. The web page then redirects the user to the hive feed page. 

In case a registered user forgets their password, the web application offers a "Forgot Password" function where the user can re-verify their identity with a verification code and reset their password and regain access to their account.

\subsubsection{Adding Hives}
Only admin users have the privilege to add, edit, and delete hives. To add a new hive an admin needs to navigate to the admin page and provide the following: 
\begin{enumerate}
    \item Hive name: A unique name to identify the hive.
    \item City: The city where the hive is located.
    \item State: The state where the hive is located.
    \item Coordinates: The geographical coordinates (latitude and longitude) of the hive's location.
    \item Picture: An image of the hive.
\end{enumerate}

Once the admin has submitted this information, a success message will be shown displayed indicating that the has been added to the list of hives in the hive feed page. However, initially, the hive will be empty, and the live data displayed will be shown as "--", indicating that no data is available and its graphs and tables will be empty. This is because there is currently no monitoring system linked to the newly added hive. Only the webmaster has the privilege of linking the monitoring system to the hive. Once the monitoring system is linked, the hive data will start to populate, and the live data, graphs, and tables will reflect the actual data collected from the hive.

\subsubsection{Hive Feed}

\begin{figure}[H]
  \includegraphics[width=\linewidth]{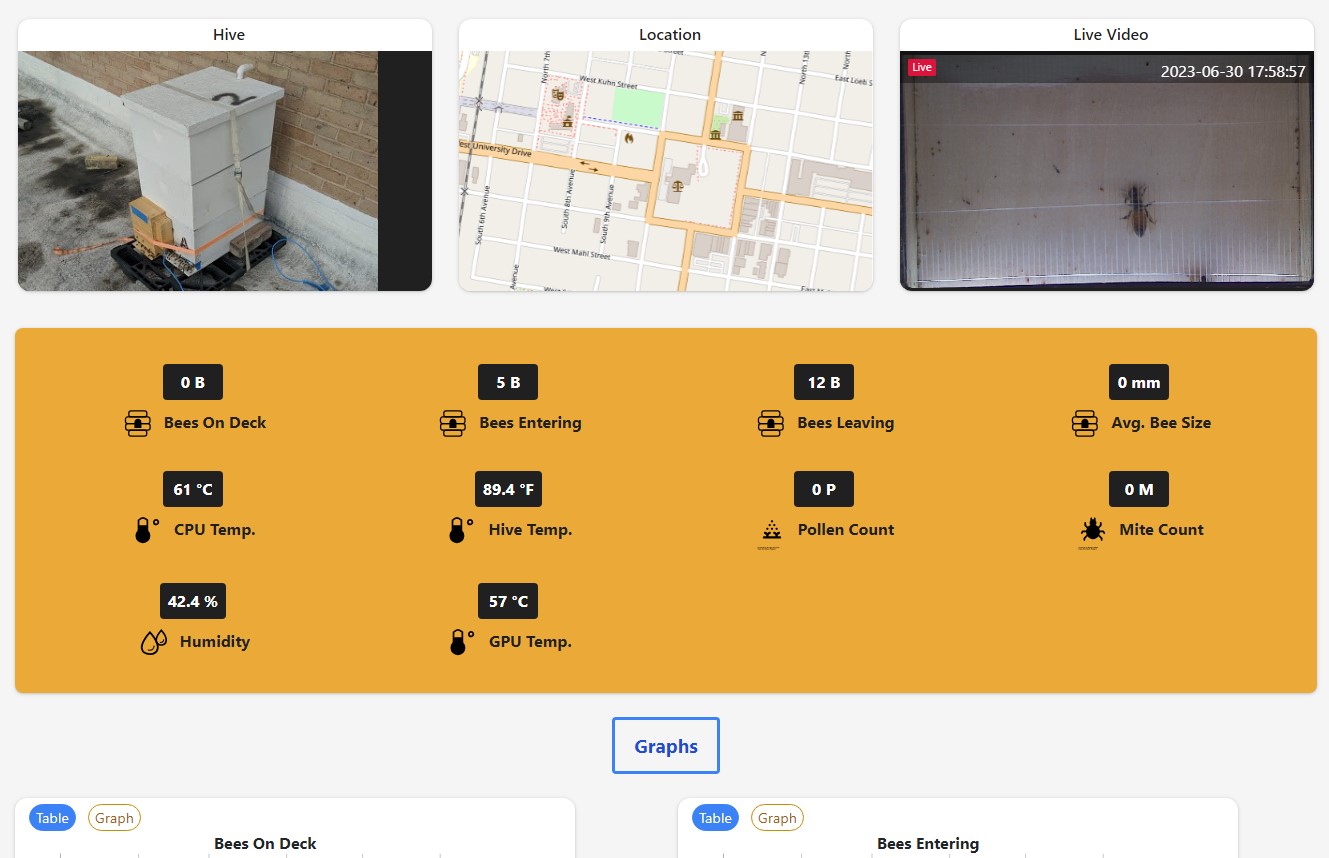}
  \caption[Honey bee hive feed users see upon logging into the website]{Honey bee hive feed users see upon logging into the website.}%
  \label{fig:IntelliDemo}
\end{figure}

Upon logging in, registered users will be directed to the hive feed page. This page showcases live and past data of each hive collected by their monitoring system. The data collected by the monitoring system is shown in Table \ref{tab:TableDataType}. On the hive feed page, the live or most recent data is displayed in the yellow block beneath the hive's image, location, and video feed, as depicted in Figure \ref{fig:IntelliDemo}. Each individual measurement is shown alongside its unit of measurement and above its title, providing a clear visualization of the data. 

The measurements are updated every 5 minutes using IntelliBeeHive's API mentioned in Section \ref{restapi}, this API facilitates communication between the web server and the user's personal device. However, it is important to note that the live video feed is available only for demo purposes and not for regular users. Regular users do not have access to a live video feed. The focus of IntelliBeeHive is to provide comprehensive data for analyzing the health of beehives, and the video feed is not considered a requirement for this analysis.

\begin{table}[H]

\renewcommand{\arraystretch}{1.3}
\caption[Data Collection and Type]{The table shows the list measurements collected from each hive to monitor their daily activity.}%
\begin{tabular}{ p{4cm} | p{4cm}}
\multicolumn{1}{c}{Measurement}
&

\multicolumn{1}{|c }{Unit of Measurement}\\
\hline

Temperature & Fahrenheit (F) \\
Humidity & Relative Humidity (\%) \\
CPU Temperature & Celsius (C) \\
GPU Temperature & Celsius (C) \\
Bees on Deck & Single Unit  \\
Bees Leaving & Single Unit  \\
Bees Arriving & Single Unit  \\
Bees Average Size & Millimeters (mm)  \\
Pollen Count & Single Unit  \\
Mite Count & Single Unit  \\

\hline
\label{tab:TableDataType}
\end{tabular}

\end{table}

\subsubsection{Graphs and Tables}\label{graphsandtables}
\begin{figure}[H]
  \includegraphics[width=\linewidth]{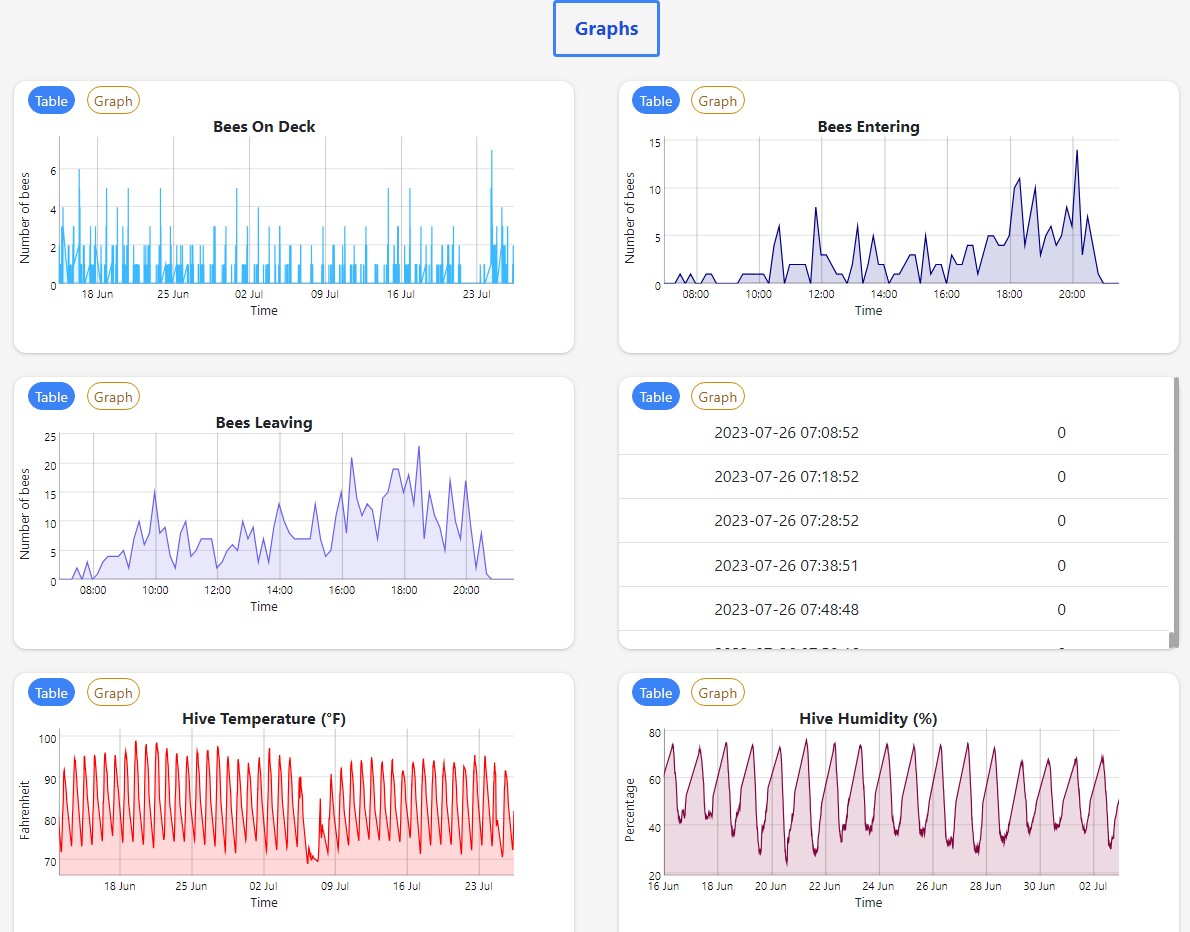}
  \caption[IntelliBeeHive Demo]{Shows 6 of 10 graphs created using Dygraphs.JS and Bootstrap libraries.}%
  \label{fig:IntelliBeeHiveGraphs}
\end{figure}

Below the yellow block containing the hive’s live measurements are a series of graphs and tables containing the past data for each measurement in Table \ref{tab:TableDataType}. There are a total of 10 blocks, one for each measurement, and users can alternate between viewing the data in graph or table format as shown in Figure \ref{fig:IntelliBeeHiveGraphs} using the 2 buttons at the top left corner of each block. 

The past data presented in these graphs and tables encompasses all the data collected from the current year, starting from January. Since hive data is uploaded every 5 minutes to the web server, a single hive can accumulate 105,120 data points for each measurement in one year. To alleviate the strain on the web server caused by loading such a large amount of data for each hive, we retrieve data collected every hour instead of every 5 minutes, significantly reducing the data size from 105,120 units per measurement to 8,760 units per measurement. This approach makes the data more manageable.

Once the data is retrieved it is rendered into table format using HTML and CSS and into graph format using Dygraphs, an open-source JavaScript charting library designed to handle large data sets \cite{dygraphs2022}. Open-source software refers to software that grants users the freedom to use, modify, and distribute the code without restrictions. How the data is retrieved will be discussed in Section \ref{Backend}.

\subsection{Backend}\label{Backend}
IntelliBeeHive is hosted on a Linux virtual machine located at the University of Texas Rio Grande Valley (UTRGV). The virtual machine serves as the web server or cloud computer for IntelliBeeHive, providing a secure and flexible environment. The web server is responsible for hosting the web application, as well as collecting, storing, and sending beehive data.

IntelliBeeHive is written in PHP, an open-source scripting language tailored for web applications, and was developed using a Laravel framework. A web framework provides an application with many useful libraries specific for web development and provides a standard structure that most web applications use. The Laravel framework is a powerful open-source framework offering numerous libraries and components for APIs and database handling and follows a standard structure that is commonly used in web applications. This section will cover IntelliBeeHive’s backend workflow, database structure, and how data is collected and sent by the API.

\subsubsection{SQL Database}\label{sqldb}
\begin{figure}[h!]
  \includegraphics[width = \linewidth]{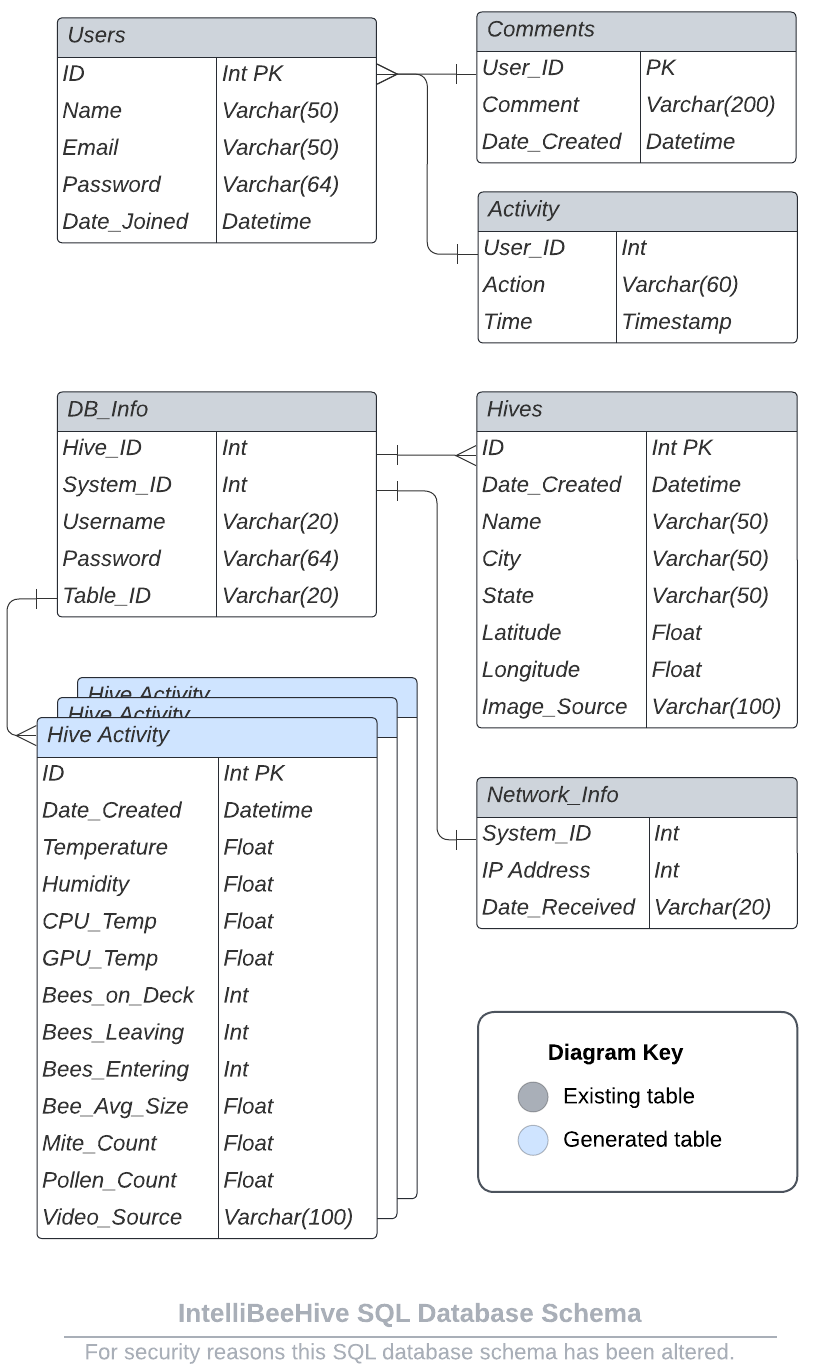}
  \caption[Database Schema]{Shows IntelliBeeHive’s SQL database schema.}%
  \label{fig:dbSchema}
\end{figure}

The IntelliBeeHive website stores all of its data in an SQL or relational database managed by MySQL, an open-source SQL management system. SQL stands for Structured Query Language and is used to create, store, update, and retrieve data from structured tables. In an SQL table, each row represents a data entry and each column identifies a specific field of the entry. 
IntelliBeeHive’s database is made up of 6 main tables: Users, Comments, Activity, Hives, DB\_Info, and Network\_Info. Figure \ref{fig:dbSchema} illustrates the logical structure of the tables. The Users, Comments, and Activity tables contain all the data pertaining to the users. The Users table contains information such as the user's name, credentials, and a primary key that uniquely identifies each user. The Comments and Activity tables store user comments and web activity respectively. These tables can be linked to a specific user through their primary key, as shown in Figure \ref{fig:dbSchema}. 
The Hives, DB\_Info, and Network tables store data pertaining to the beehives. The Hives table stores a hive’s name, location, picture, and primary key, and the Network\_Info table stores the hive’s monitoring system’s identification key. Whenever a new monitoring system is assigned or added to a hive by the webmaster, a new Hive Activity table is created with a unique title, serving as a key. Each Hive Activity table stores the measurements listed in Table \ref{tab:TableDataType} for a specific hive. Thus, there is a separate Hive Activity table for each hive in the system. The DB\_Info table stores a hive’s primary key, system identification key, and table key to link each hive to their Hive Data table and monitoring system.

\subsubsection{Backend Workflow}
IntelliBeeHive’s back-end workflow is similar to its front-end workflow covered in Section \ref{layout}, however in this section we will discuss the underlying processes.

When a user visits the Landing Page, they have several options: they can view the Hive Feed Demo page, create a new account through the Sign Up page, or log into their existing account. If a user opens the Hive Demo page, the hive data is fetched from the SQL database using the API. Since hive activity data will be continuously sent to the user's browser from the web server every 5 minutes, the API is used to facilitate this process. On the other hand, when a user creates an account through the Sign Up page, their input information is submitted to the web server without the use of the API. The API is primarily reserved for scenarios where data needs to be frequently sent from or received by the web server. If the submitted information is correct, the user is assigned a token, which serves as a verification of their access and privileges. Subsequently, they are redirected to the Hive Feed page. If the information is incorrect the user is sent back to the Sign Up page.

Similarly, when a user logs into the application their credentials will be queried and verified against the stored information in the SQL database. If the credentials exist and match then the application will determine if the user should have admin privileges. If the user is an admin, they will be assigned a special token that identifies them as an admin and redirects them to the Admin Page, else they’ll be assigned a regular token and redirected to the Hive Feed page. The Hive Feed page similar to the Hive Feed Demo page uses the API to fetch all hive past and current activity data.

\begin{figure}[H]
\centering
  \includegraphics[width = \linewidth]{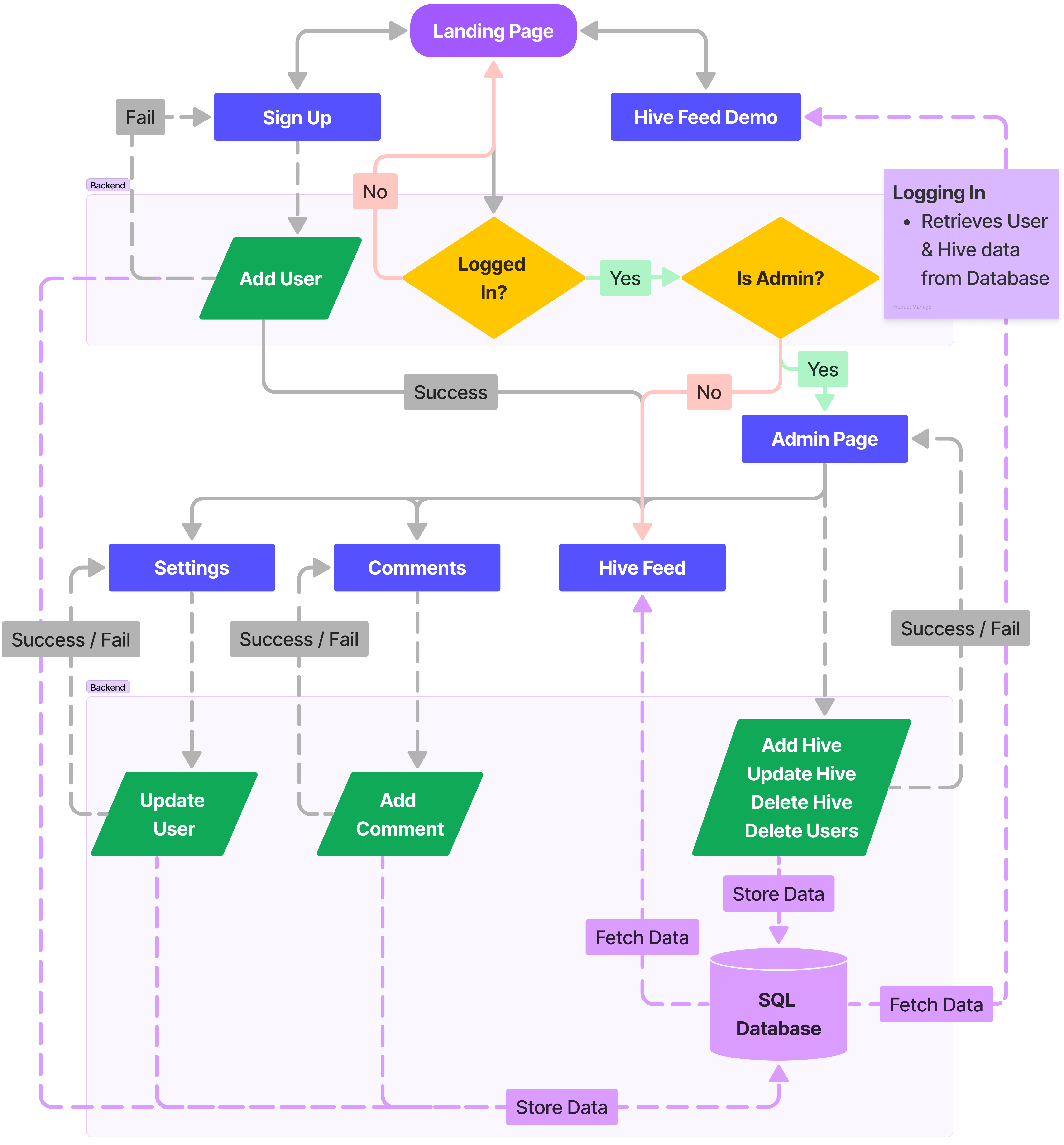}
  \caption[IntelliBeeHive's backend workflow]{Shows a flowchart diagram of IntelliBeeHive’s backend workflow.}%
  \label{fig:IntelliBeeHiveBackendWorkflow}
\end{figure}

\subsubsection{Adding Users, Activities, Comments and Hives}
Once a user is logged in they can add comments, update their credentials, or manage their hives. Regular users can add comments and update or delete their credentials, meanwhile admin users can do the same plus add, update, and delete hives. 

\begin{figure}[h!]
  \includegraphics[width = \linewidth]{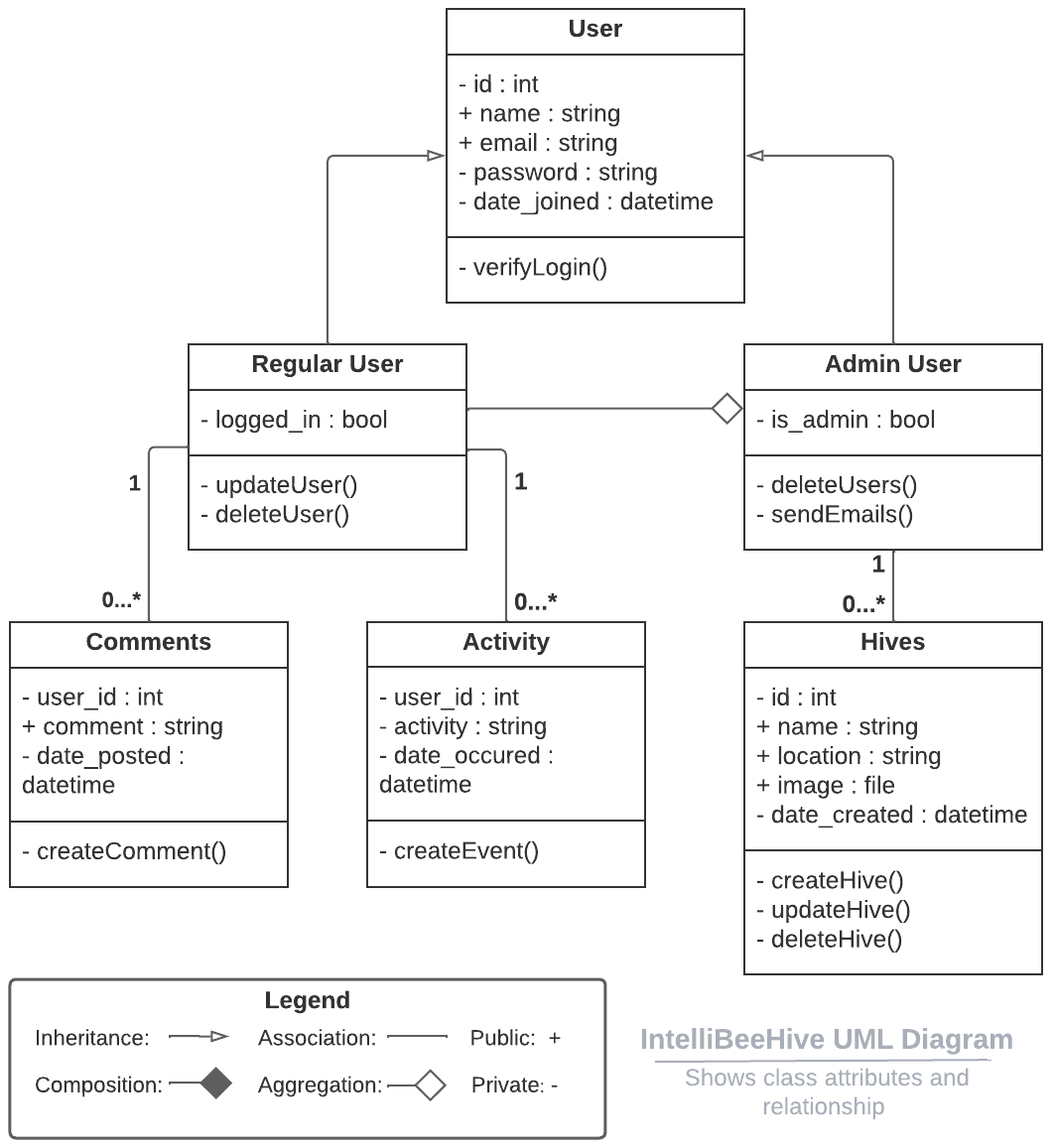}
  \caption[IntelliBeeHive's UML diagram]{Shows IntelliBeeHive’s UML diagram.}%
  \label{fig:IntelliBeeHiveUML}
\end{figure}

We can consider each user, comment, activity, and hive as a class with its own set of attributes mentioned in Section \ref{sqldb}. An instance of a class can be considered an object. For example, when an action is performed, an instance of the corresponding class is created, which can be seen as an object. We can use a UML (Unified Modeling Language) diagram to represent the relationship and interaction between these classes. Figure \ref{fig:IntelliBeeHiveUML} shows a UML diagram of our user, comment, activity, and hive classes. Each box in the UML diagram represents an object and is made up of 3 sections, going from top to bottom: class name, list of attributes, and list of privileges. Attributes input by the user are marked as public (+) and must be valid, else an object is not created and the user is sent a fail message. 
A regular and admin user are objects inherited from the user class since they both have the same attributes but differ in privileges. An admin user is an aggregation of a regular user since it has the privileges of a regular user in addition to its own. A regular user can create multiple comment and activity objects that will be associated with the user who created them by their primary key.  However, unlike comments and activity objects, when a hive object is created there is no key associating the hive to who created it. The only association the hive object has with the admin user is that only admin users can create hives. When any object is created they are stored in the SQL database. Hive, comment, and activity objects will continue to exist without the user who created them, thus why they are only associated with the user.

\subsubsection{REST API}\label{restapi}
IntelliBeeHive's API follows a REST (Representational State Transfer) architecture, which adheres to several design principles. These principles include having a uniform interface, separating the client and server, being stateless, and employing a layered system architecture \cite{restapiarch2017}. A uniform interface means every request made to the API should work the same. The client and server refer to two separate computers, one making the request and the other fulfilling the request. In our case, the computer making the request is either the monitoring system or the web browser, and the computer fulfilling the request is the web server. The requests must be stateless, meaning each request should have the necessary information for the web server to fulfill without the need for a second request. The life cycle of a request follows a layered system architecture. The client layer handles sending requests and receiving responses from the API that includes a status code that indicates whether the request succeeded or failed. The authentication layer verifies if the client is authorized to access the API, for authorization the client must provide an alpha-numeric authentication key. The endpoint layer verifies if the client’s input data is valid and formats the request’s output data in JSON, a lightweight data-interchange format. The data access layer is responsible for handling the client's input data by checking for and removing any malicious code, preparing the necessary database query to retrieve or store data, and determining the success of the query execution. The database layer executes the query and returns the output to the data access layer, this layer occurs in MySQL which is covered in Section \ref{sqldb}.

\begin{figure}[H]
\centering
  \includegraphics[width = \linewidth]{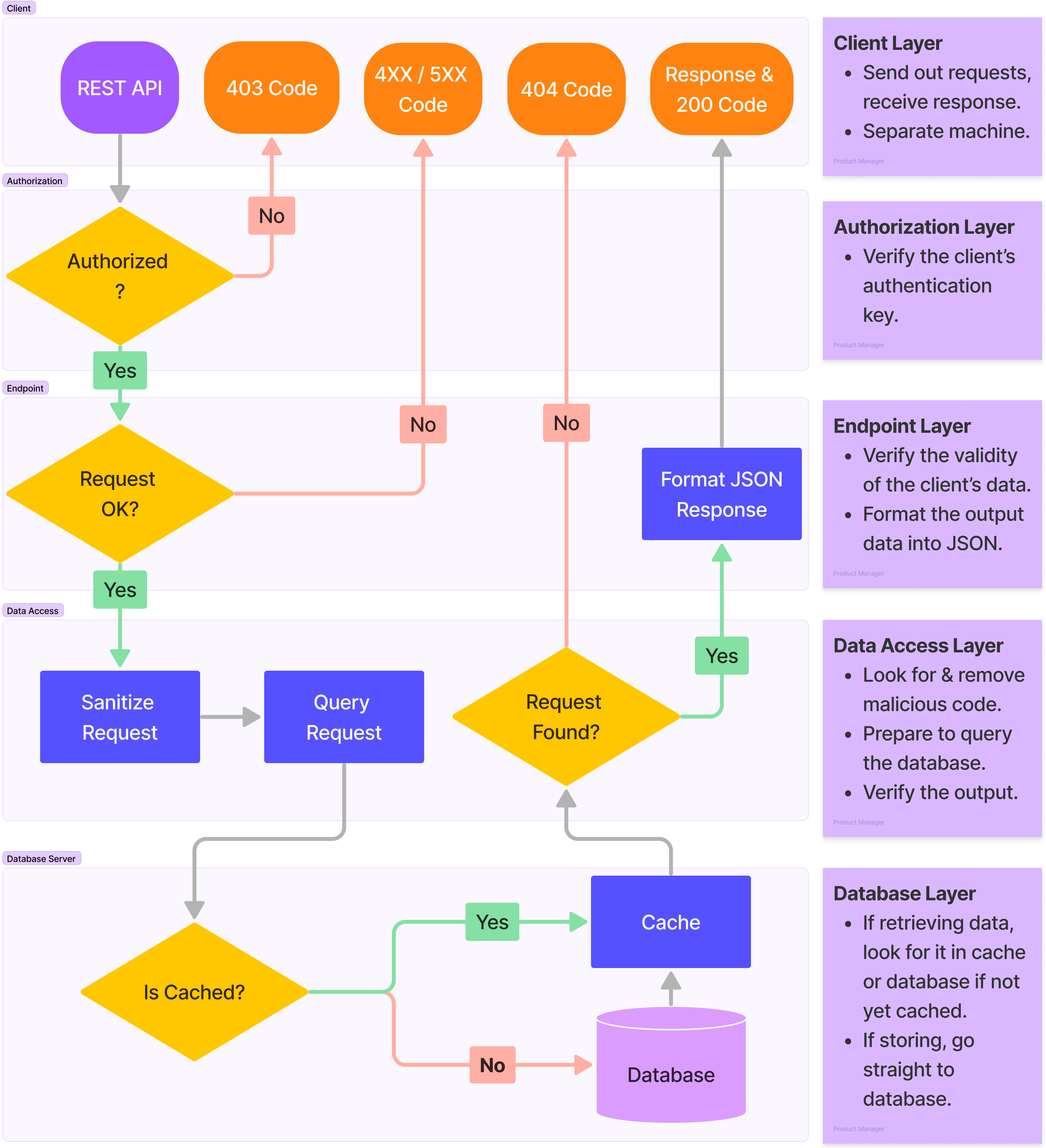}
  \caption[IntelliBeeHive's REST API]{Shows a flowchart diagram of IntelliBeeHive’s REST API request workflow.}%
  \label{fig:IntelliBeeHiveRESTAPI}
\end{figure}

\subsubsection{Collecting and Retrieving Honey Bee Data}
\begin{figure}[h!]
  \includegraphics[width = \linewidth]{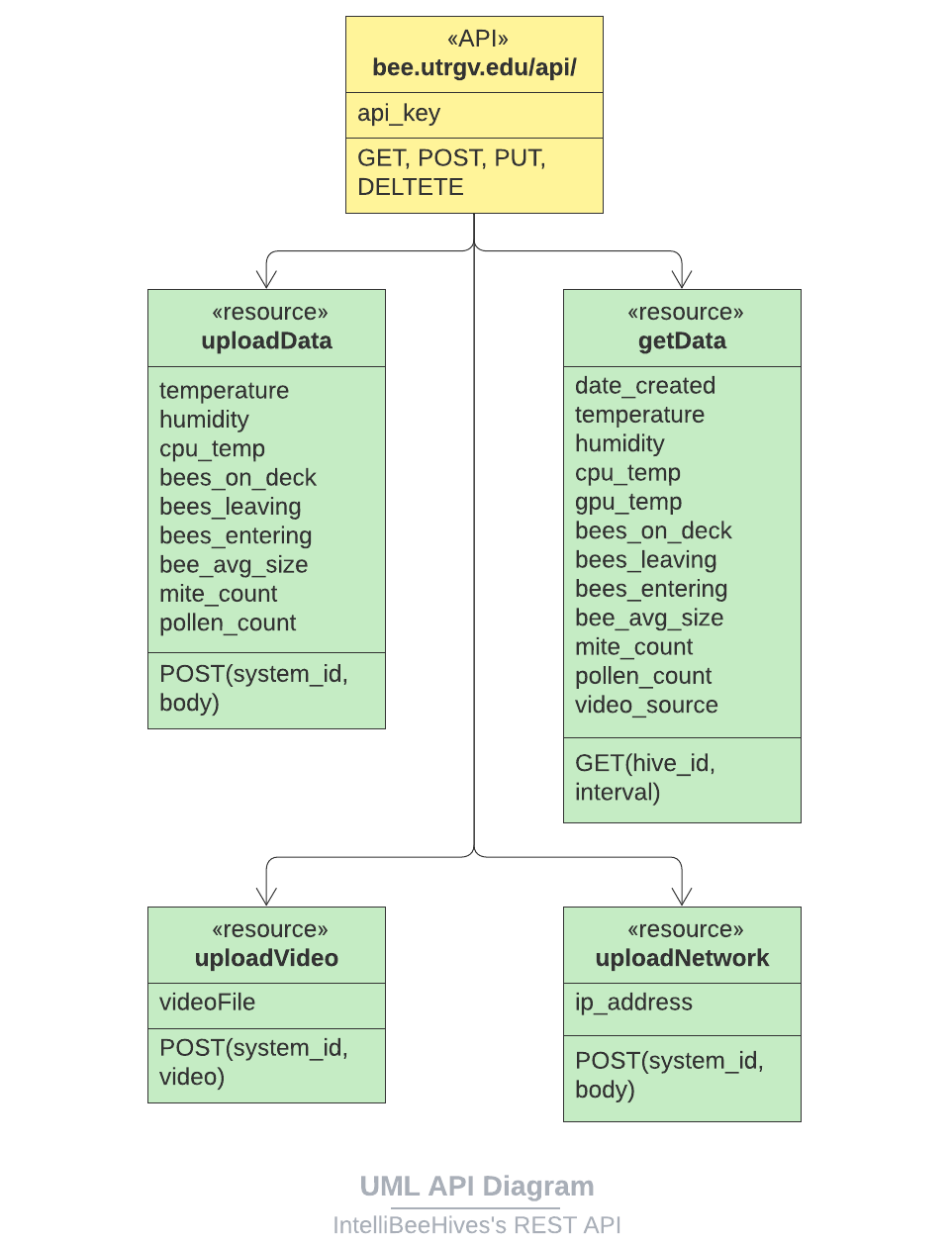}
  \caption[IntelliBeeHive's REST API UML diagram]{Shows a UML diagram of IntelliBeeHive’s REST API.}%
  \label{fig:IntelliBeeHiveRESTAPIUML}
\end{figure}

IntelliBeeHive’s REST API has has the following 4 main operations: getData, uploadData, uploadVideo, and uploadNetwork. The UML diagram in Figure \ref{fig:IntelliBeeHiveRESTAPIUML} depicts each operation in blocks. Each block is made up of 3 parts, going from top to bottom: the purpose and URL of the operation, the variable data being sent/received, and the REST API request type. Three of the operations are of type POST and are used only by the monitoring system. POST requests in a REST API are used to upload data, thus they are used exclusively by the monitoring system to upload the hive's environment condition, video feed of the hive, and network information of the system. On the other hand, GET requests in an API are used to retrieve data and thus are used by the website's Hive Feed and Hive Feed Demo pages to display the hive's latest condition and video feed. Although the REST API and the website are hosted on the same machine, the GET request is made from the user's browser located on a different machine. The reason behind making GET requests to the API from the user's machine is to give the user live updates without them having to refresh their browser. When a user opens up a page to any website they receive a static page that won't change unless they re-query the web server by refreshing their browser. Our page contains a JavaScript script that queries the web server using the REST API to provide the user with the newest updates every 5 minutes without them having to refresh their browser.

\section{Results}\label{ch:Results}
\subsection{YOLOV7 Training}

\begin{figure}[H]
    \centering
    \begin{subfigure}[b]{0.24\textwidth}
        \centering
        \includegraphics[width=\linewidth]{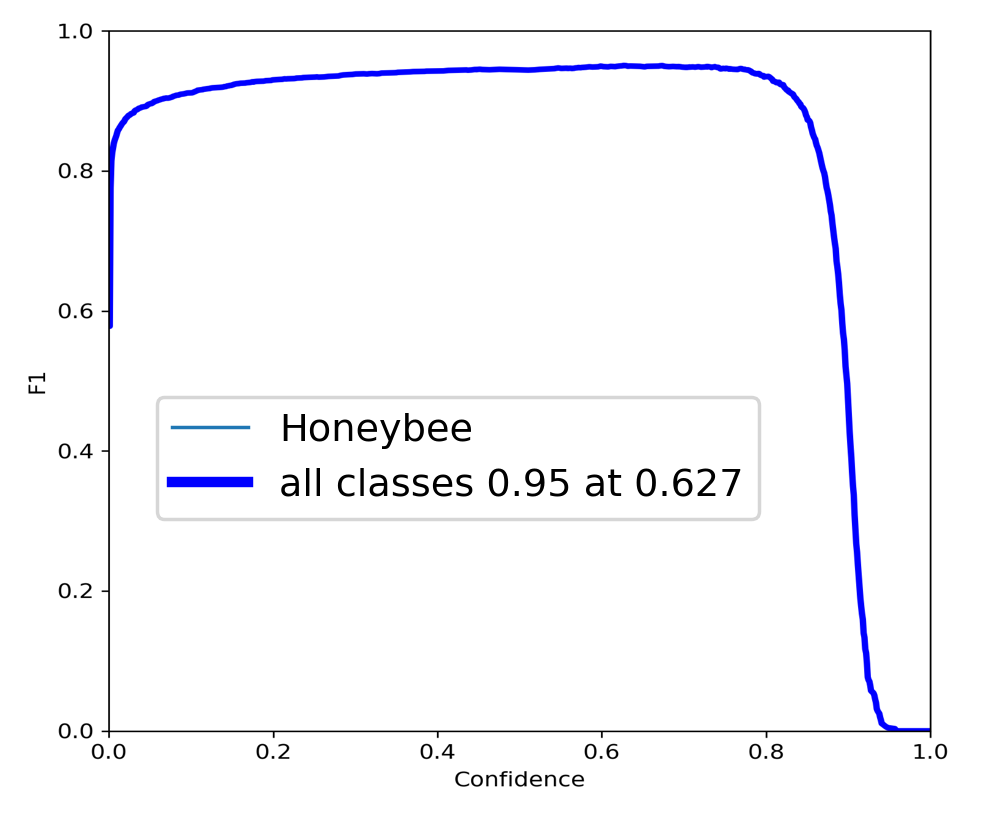}
        \caption[Honey Bee F1 Curve]{F1 curve for honey bee object detection model.}%
        \label{fig:HoneyBee_F1_curve}
    \end{subfigure}
    \hfill
    \begin{subfigure}[b]{0.24\textwidth}
        \centering
        \includegraphics[width=\linewidth]{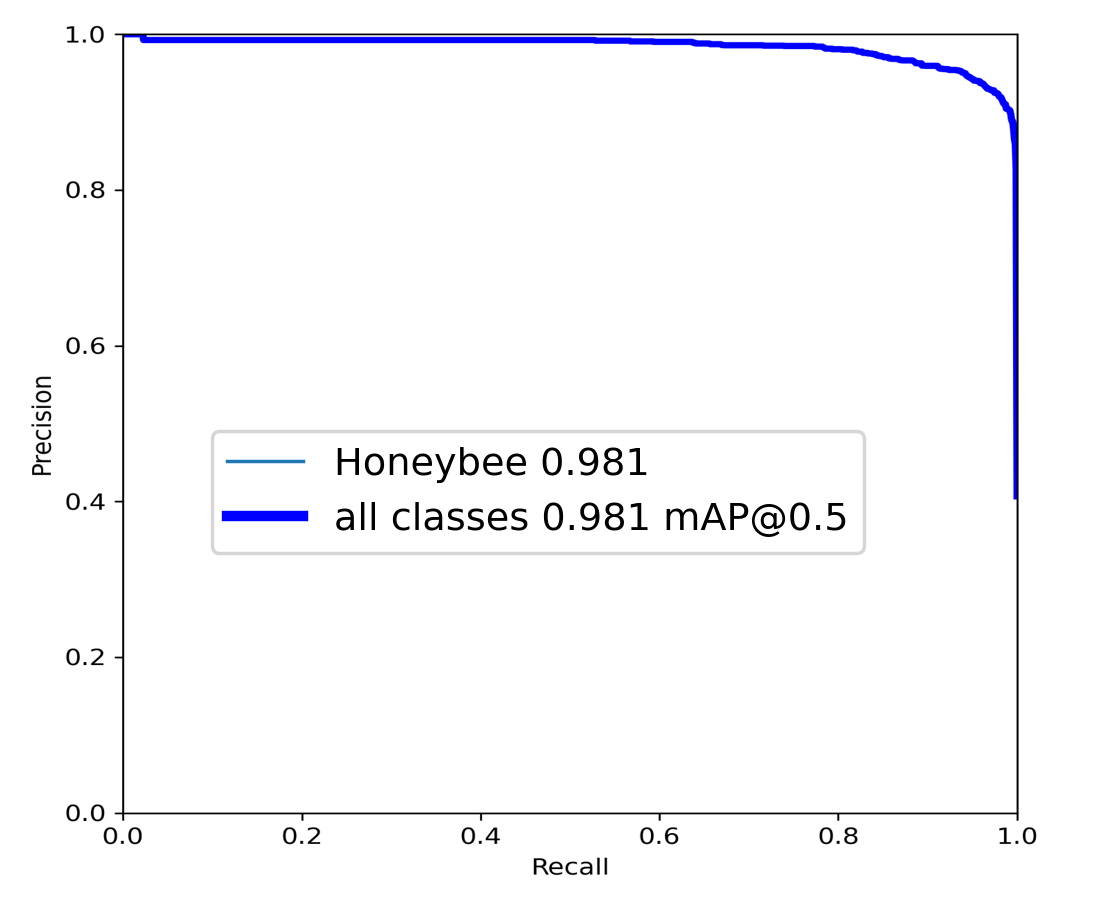}
        \caption[Honey Bee PR Curve]{Precision and Recall curve for honey bee object detection model.}%
        \label{fig:HoneyBee_PR_curve}
    \end{subfigure}
    \caption[Honey Bee Model Training Results]{Honey bee model training results}
\end{figure}

The graphs shown below are the results of our YOLOv7-Tiny model's training. The F1-score for honey bee model recognition
is 0.95 and the precision and recall value is 0.981 as shown in Figure \ref{fig:HoneyBee_F1_curve} and \ref{fig:HoneyBee_PR_curve}.

\begin{figure}[H]
    \centering
    \begin{subfigure}[b]{0.24\textwidth}
        \centering
        \includegraphics[width=\linewidth]{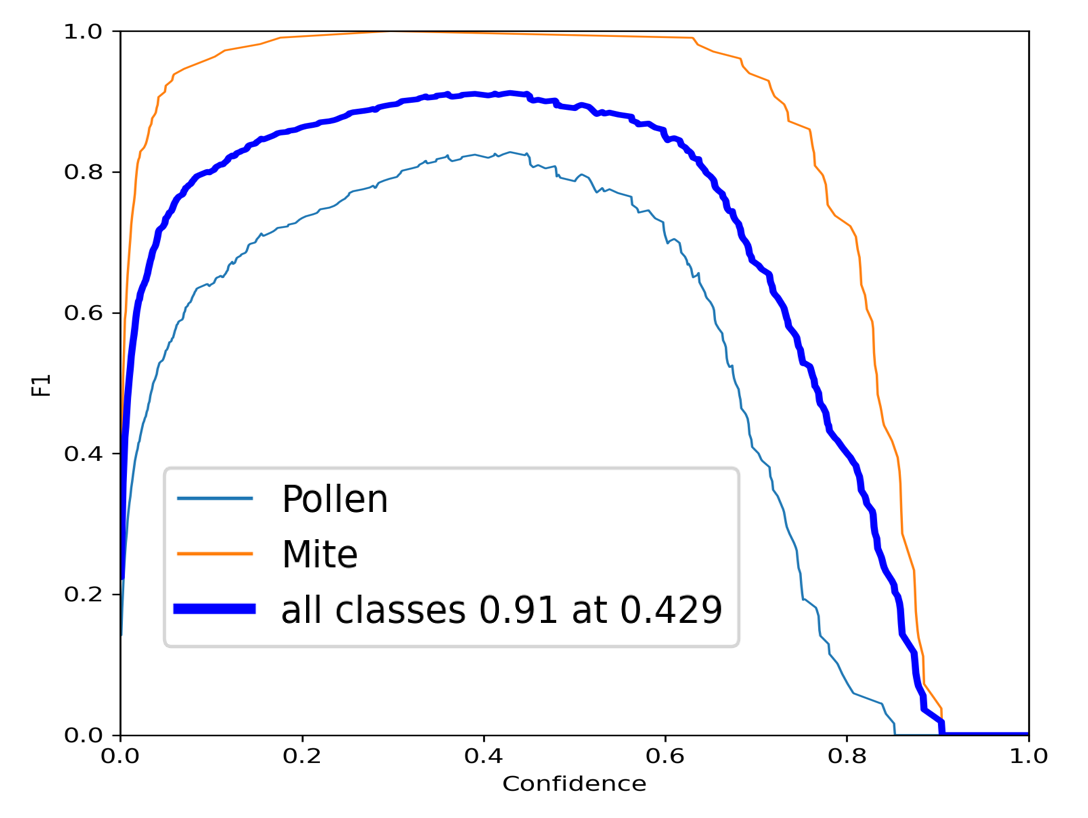}
        \caption[Pollen and Mite F1 Curve]{F1 curve for pollen and mite object detection model.}%
        \label{fig:PolMite_F1_curve}
    \end{subfigure}
    \hfill
    \begin{subfigure}[b]{0.24\textwidth}
         \centering
        \includegraphics[width=\linewidth]{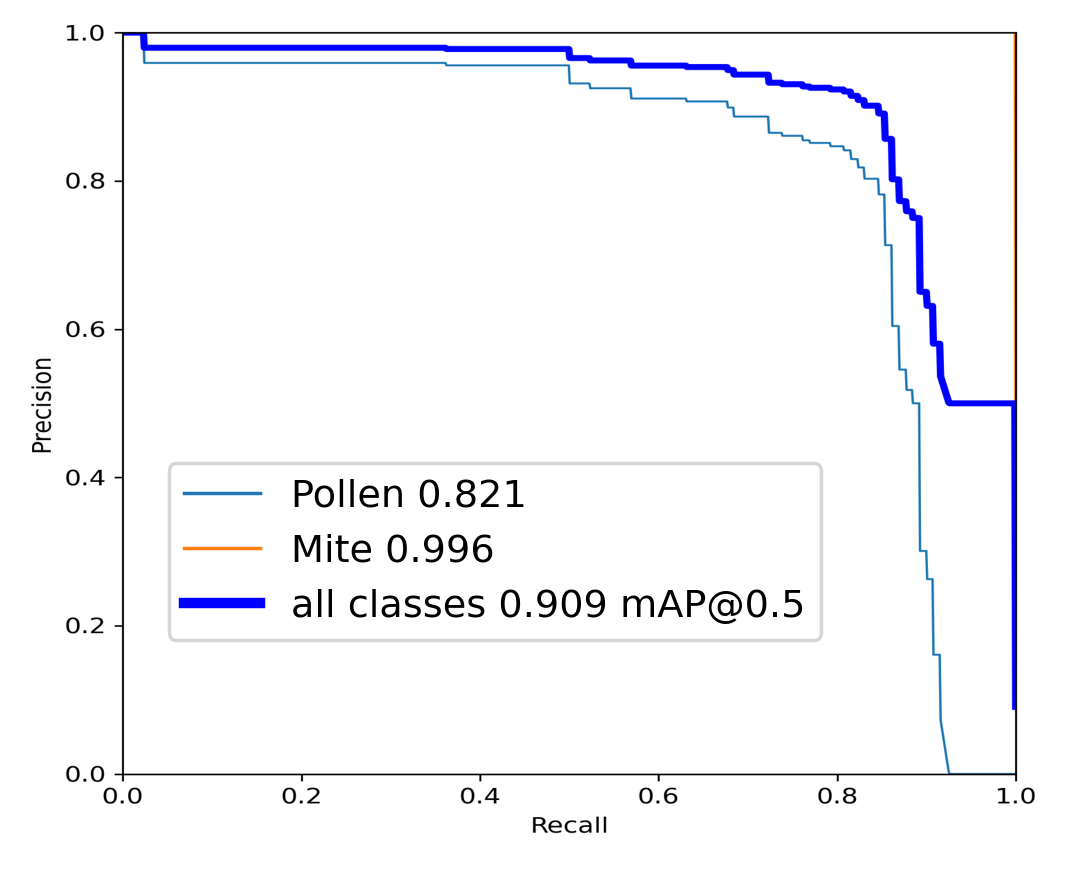}
        \caption[Pollen and Mite PR Curve]{Precision and Recall curve for  pollen and mite object detection model.}%
        \label{fig:PolMite_PR_curve}
    \end{subfigure}
    \caption[Pollen and Mite Model Training Results]{Pollen and Mite model training results}
\end{figure}

For our pollen and mite object detection model F1-score is 0.95 and the precision and recall value is 0.821 for pollen and 0.996 for mite as shown in Figure \ref{fig:PolMite_F1_curve} and \ref{fig:PolMite_PR_curve}.

The images shown below are extracted frames from video output after it's processed by the honey bee YOLOv7 tiny model and our tracking algorithm. The circle around each detection is the freedom where the honey bee can move and still be considered the same honey bee. The blue dot represents the honey bees' previous mid-point and the red dot represents the current mid-point.

\begin{figure*}[h!]
    \centering
    \begin{subfigure}{.45\textwidth}
        \includegraphics[width=\linewidth]{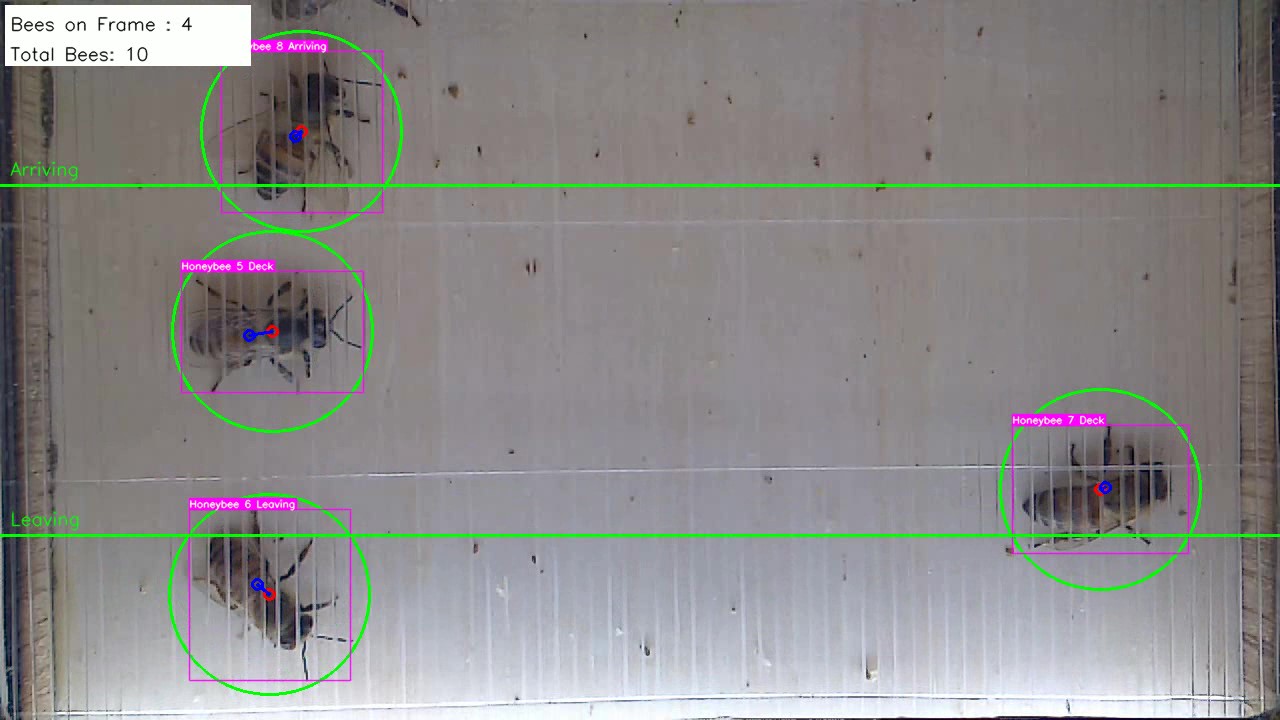}
        \caption{Honey Bee Video Frame Tracking Example 1}\label{fig:FrameExample1}
    \end{subfigure}
    \begin{subfigure}{.45\textwidth}
        \includegraphics[width=\linewidth]{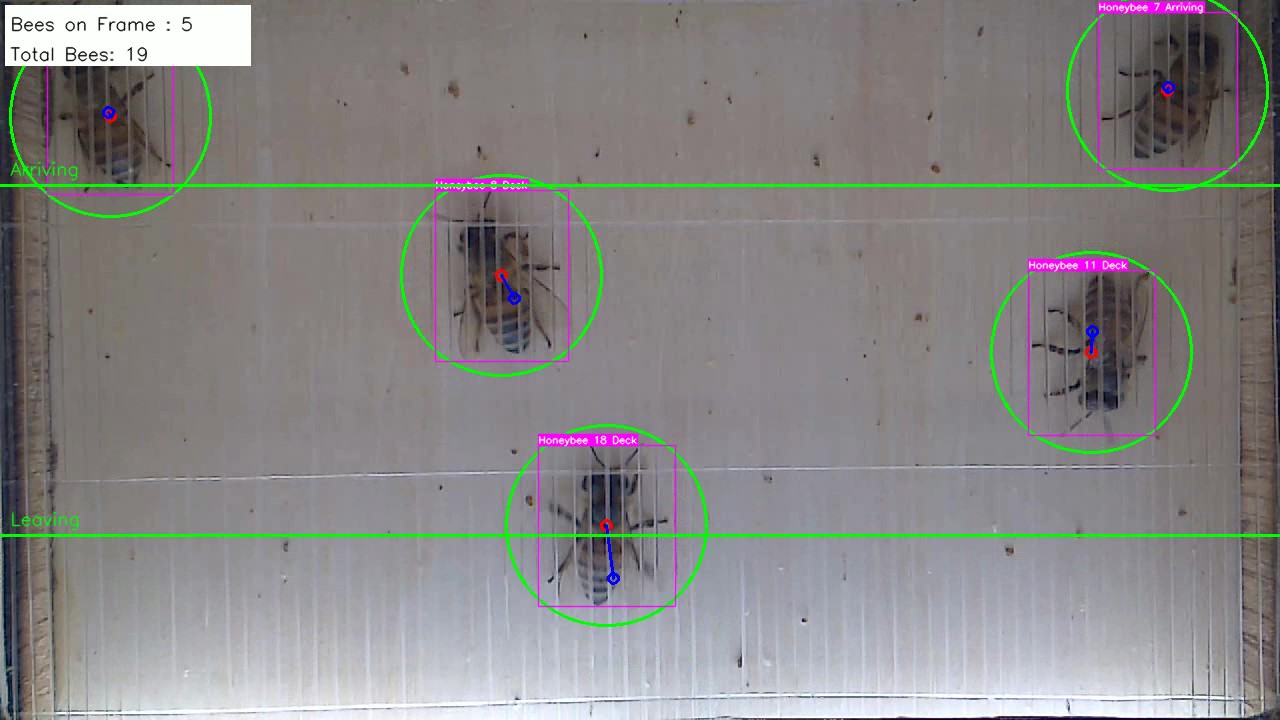}
        \caption{Honey Bee Video Frame Tracking Example 2}\label{fig:FrameExample2}
    \end{subfigure}
    
    \begin{subfigure}{.45\textwidth}
        \includegraphics[width=\linewidth]{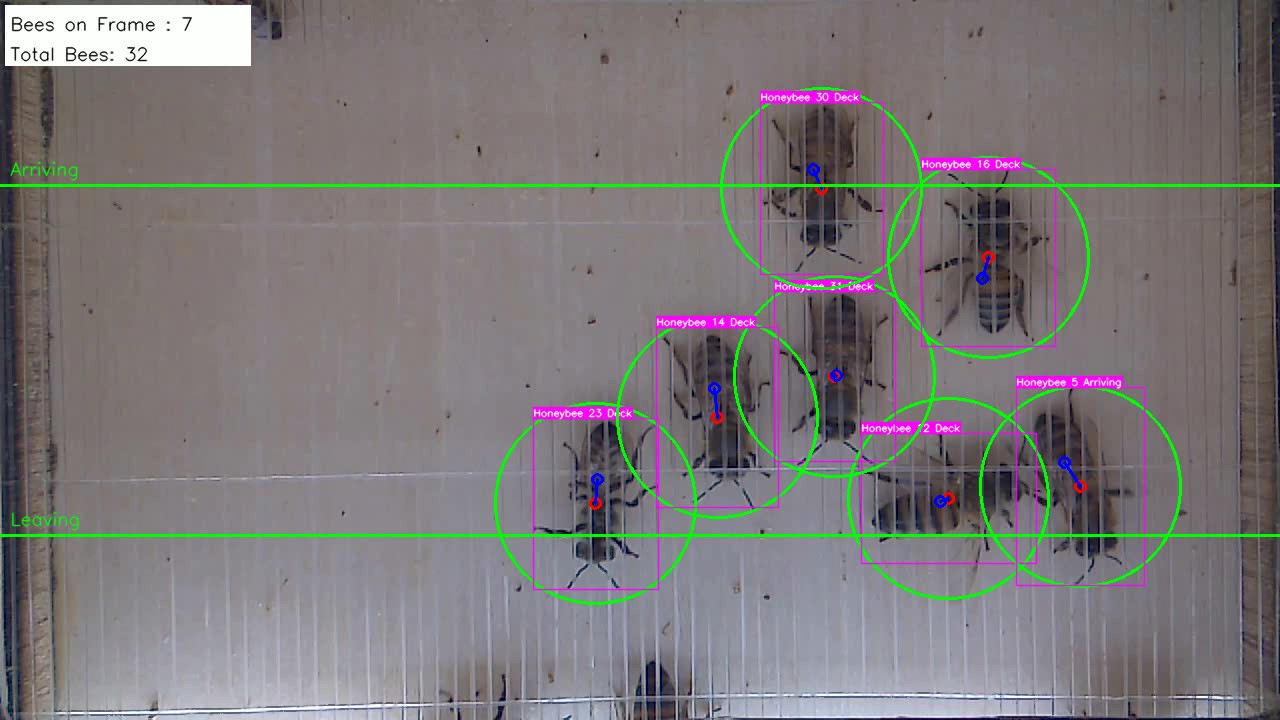}
        \caption{Honey Bee Video Frame Tracking Example 3}\label{fig:FrameExample3}
    \end{subfigure}
    \centering
    \begin{subfigure}{.45\textwidth}
        \includegraphics[width=\linewidth]{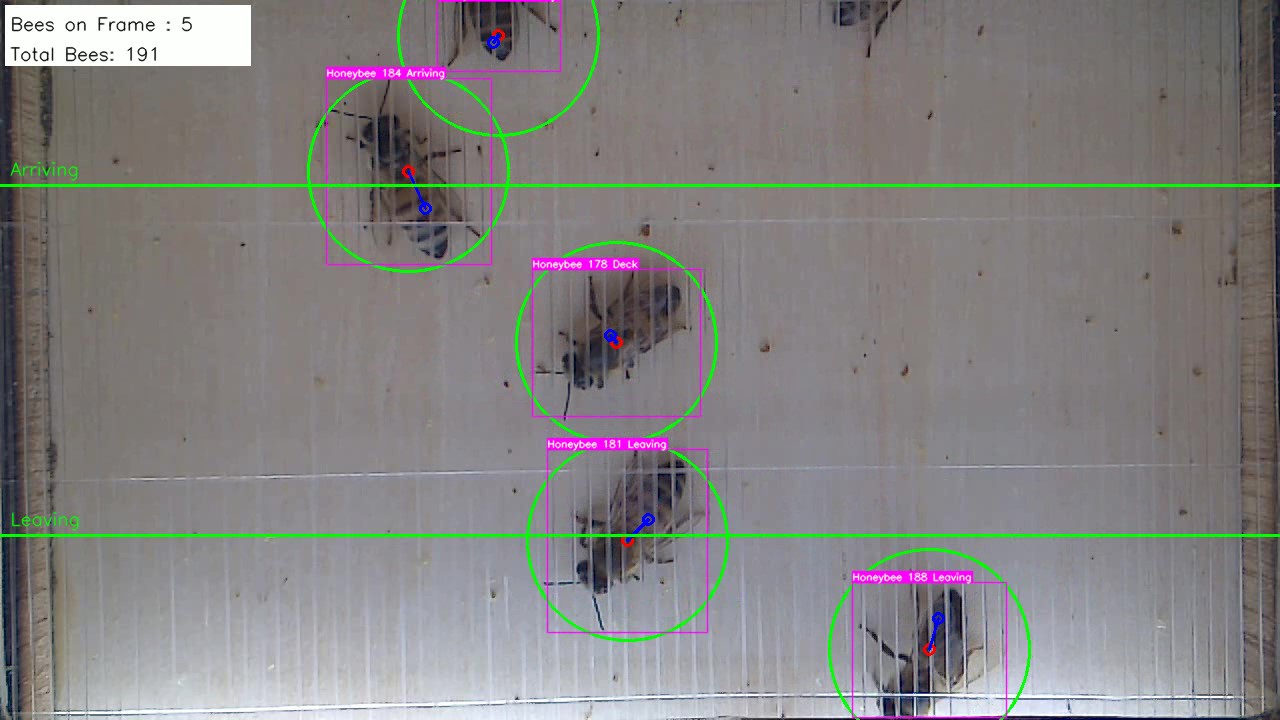}
        \caption{Honey Bee Video Frame Tracking Example 4}\label{fig:FrameExample4}
    \end{subfigure}
    \caption[Honey Bee Tracking Output Example]{Honey Bee Tracking Output Example}
\end{figure*}

 The figures below are example outputs of our pollen and mites detection model using our TensorRT engine on each honey bee. The letter P indicates pollen was detected followed by the confidence of the model. 

\begin{figure}[H]
    \centering
    \begin{subfigure}[b]{0.22\textwidth}
    \centering
        \includegraphics[scale=0.7]{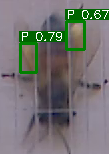}
  \caption[Honey Bee Pollen]{Honey bee Example 1 with Pollen}%
    \label{fig:HoneybeePollen1}
    \end{subfigure}
    \hfill
    \begin{subfigure}[b]{0.22\textwidth}
    \centering
       \includegraphics[scale=0.7]{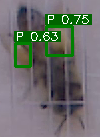}
  \caption[Honey Bee Pollen]{Honey bee Example 2 with Pollen}%
    \label{fig:HoneybeePollen2}
    \end{subfigure}
    \begin{subfigure}[b]{0.22\textwidth}
    \centering
  \includegraphics[scale=0.7]{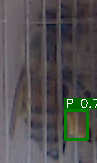}
  \caption[Honey Bee Pollen]{Honey bee Example 3 with Pollen}%
    \label{fig:HoneybeePollen9}
    \end{subfigure}
    \hfill
    \begin{subfigure}[b]{0.22\textwidth}
    \centering
  \includegraphics[scale=0.7]{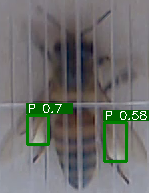}
  \caption[Honey Bee Pollen]{Honey bee Example 4 with Pollen}%
    \label{fig:HoneybeePollen4}
    \end{subfigure}
    \begin{subfigure}[b]{0.22\textwidth}
    \centering
  \includegraphics[scale=0.7]{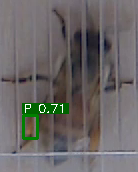}
  \caption[Honey Bee Pollen]{Honey bee Example 5 with Pollen}%
    \label{fig:HoneybeePollen5}
    \end{subfigure}
    \hfill
    \begin{subfigure}[b]{0.22\textwidth}
    \centering
  \includegraphics[scale=0.7]{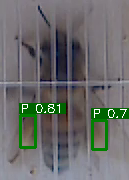}
  \caption[Honey Bee Pollen]{Honey bee Example 6 with Pollen}%
    \label{fig:HoneybeePollen6}
    \end{subfigure}
    \caption[Pollen Detection Output]{Pollen Detection Example Output Images}
\end{figure}

Due to our "mite" detection model being trained with placeholder data, we will not go in-depth into our model's accuracy in detecting mites. 

\begin{figure}[H]
    \centering
  \includegraphics[scale=0.7]{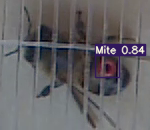}
  \caption[Honey Bee Mite]{Honey bee Example 1 with Mite}%
    \label{fig:HoneyBeeMite}
\end{figure}

\subsection{Ground Truth Data vs Tracking Algorithm}

To evaluate the accuracy of our algorithm, we conducted an experiment using five 1-minute long videos. Each video was manually labeled tracking each honey bee's identification, final status, initial frame detected, and last frame seen. We processed the videos through our algorithm to obtain the algorithm's output. The results for the five videos are presented in Table \ref{tab:trackingResult}.

\begin{table}[ht]
\renewcommand{\arraystretch}{1.5}
\caption[Tracking Algorithm]{This table shows a performance comparison between our manual (M) and algorithm (A) output}%
\begin{tabular}{ p{.6cm} | p{.4cm} p{.4cm} | p{.4cm} p{.4cm} | p{.3cm} p{.3cm} | p{.4cm} p{.4cm} | p{.3cm} p{.3cm}}

\multicolumn{1}{c}{}
&
\multicolumn{2}{|c}{Arriving}
&

\multicolumn{2}{|c}{Leaving}
&

\multicolumn{2}{|c}{Deck}\
&
\multicolumn{2}{|c}{Total}\
&

\multicolumn{2}{|c}{Pollen}\\
\hline

Vid & M & A & M & A & M & A & M & A  & M & A \\
\hline

1 & 17 &17 & 19 & 19 & 0 & 0  & 36 & 36 &2 & 1\\
2 & 36 &39 & 32 & 29 & 3 & 4 & 71 & 72 &1 & 1\\
3 & 44 &42 & 34 & 33 & 1 & 4 & 79 & 79 &0 & 0\\
4 & 33 &35 & 22 & 22 & 0 & 5 & 55 & 62 &0 & 0\\
5 & 40 &40 & 34 & 42 & 1 & 7 & 75 & 79 &2 & 1\\

\hline
\end{tabular}
\label{tab:trackingResult}

\end{table}

We determine the accuracy of our algorithm by extracting the error rate using the number of "Arriving" and "Leaving" counts of honey bee's status given by the algorithm ($C_{\text{Algorithm}}$) compared to the manual count ($C_{\text{Manuel}}$) using the Equation \ref{equ:ErrorRate} below. 

\begin{equation}
\text{Error Rate} = \frac{{\left| C_{\text{Algorithm}} - C_{\text{Manuel}} \right|}}{{C_{\text{Manuel}}}}
\label{equ:ErrorRate}
\end{equation}

Once we have the Error Rate of our Algorithm we can then extract the accuracy by using Equation \ref{equ:Accuracy}.
\begin{equation}
\text{Accuracy} = 1 - {Error Rate}
\label{equ:Accuracy}
\end{equation}

We calculate the average accuracy for each video and then calculate the overall accuracy across all 5 videos to determine the accuracy of our tracking algorithm and honey bee object detection model.

Formula Key: Error = Error Rate, Arr = Arriving, Acc = Accuracy

\begin{align*}
\text{Error}_1 &= \text{Arr} \ \frac{17 - 17}{17} = 1.0000    \ \ \ \text{Leaving}\frac{19 - 19}{19} = 1.0000 \\
\text{Acc}_1 &= 1 - \text{Error}_1 = 1 - \frac{1.0000 + 1.0000}{2} = 1.0000 \\
\\
\text{Error}_2 &= \text{Arr} \ \frac{39 - 36}{36} = 0.9166    \ \ \ \text{Leaving}\frac{29 - 32}{32} = 0.9062 \\ 
\text{Acc}_2 &= 1 - \text{Error}_2 = 1 - \frac{0.9166  + 0.9062}{2} = 0.9114 \\
\\
\text{Error}_3 &= \text{Arr} \ \frac{42 - 44}{44} = 0.9545    \ \ \ \text{Leaving}\frac{33 - 34}{34} = 0.9705 \\
\text{Acc}_3 &= 1 - \text{Error}_3 = 1 - \frac{0.9545 + 0.9705}{2} = 0.9625 \\
\\
\text{Error}_4 &= \text{Arr} \ \frac{35 - 33}{33} = 0.9393 \ \ \text{Leaving}\frac{22 - 22}{22} = 1.0000 \\
\text{Acc}_4 &= 1 - \text{Error}_4 = 1 - \frac{0.9393  + 1.0000}{2} = 0.9696 \\
\\
\end{align*}
\begin{align*}
\text{Error}_5 &= \text{Arr} \ \frac{40 - 40}{40} = 1.0000    \ \ \ \text{Leaving}\frac{32 - 34}{34} = 0.941 \\
\text{Acc}_5 &= 1 - \text{Error}_5 = 1 - \frac{1 + 0.9411}{2} = 0.9705 \\
\\
\text{Avg Acc} &= \frac{\text{Acc}_1 + \text{Acc}_2 + \text{Acc}_3 + \text{Acc}_4 + \text{Acc}_5}{5} \\
&= \frac{1.0000 + 0.9114 + 0.9625 + 0.9696 + 0.9705}{5} \\
&\approx 0.9628 \quad \text{(or } 96.28\% \text{)}
\end{align*}

We exclude honey bees with a "New" status from our analysis due to the potential unreliability of their count. This is because honey bees have the ability to stay near the entrance and exit of the container, which can create complications for the model in accurately determining whether an object is indeed a honey bee or not.

The "Deck" difference happens due to our approach in our algorithm. The issue arises when the algorithm relies on identifying the nearest honey bee in each frame to track their movement. However, if a honey bee happens to move significantly faster than usual, this approach can lead to problems. Specifically, when the algorithm considers the closest midpoint in the next frame as the same bee, it may result in losing track of the current honey bee and mistakenly pairing other honey bees with the wrong counterparts. This can lead to unpaired honey bees being marked as new and potentially disrupting the tracking process. Increasing the frame rate can significantly improve this problem.

To measure the accuracy of our pollen and mite detection, because the five 1-minute videos do not give us enough honey bees with pollen as shown in Table \ref{tab:trackingResult}, we manually annotated honey bee profile images only for five different 5-minute videos shown in Table \ref{tab:pollenResult}. The pollen model results include the counts of false positives and false negatives, as well as the total number of honey bees detected for each video. Due to our limitation on mite data, we aren't able to accurately represent the accuracy of our mite detection class. 
\\
\begin{table}[ht]
\renewcommand{\arraystretch}{1.3}
\caption[Pollen Model]{This table shows the performance of our pollen model where M is the Manually counted total of honey bees with pollen and A is the Algorithms total count of honey bees with pollen.}%
\begin{tabular}{ p{.6cm} | p{.3cm} p{.3cm} | p{1.5cm} | p{1.5cm} | p{1.5cm}}

\multicolumn{1}{c}{}
&

\multicolumn{2}{|c}{Pollen}\
&
\multicolumn{1}{|c}{}\
&
\multicolumn{1}{c}{}\
&
\multicolumn{1}{c}{}\\
\hline

Vid & M & A & False Pos. & False Neg. & Total Bees\\
\hline

\centering 1 &23 & 22 & \centering 3 & \centering 4 & 325\\ 
\centering 2 &21 & 14 & \centering 1 & \centering 8 & 296\\
\centering 3 &10 & 6 & \centering 1 & \centering 4 & 267\\ 
\centering 4 &7 & 7 & \centering 0 & \centering 0 & 209\\ 
\centering 5 &15 & 15 & \centering 2 & \centering 2 & 253\\ 

\hline
\end{tabular}
\label{tab:pollenResult}

\end{table}

To determine the accuracy of our pollen detection model we use the Precision \ref{equ:PrecisionFormula} and Recall \ref{equ:RecallFormula} formulas to then extract our F1 scores \ref{equ:F1ScoreFormula}.

\begin{equation}
\text{Precision} = \frac{{\text{True Positive}}}{\text{True Positive} + \text{False Positive}}
\label{equ:PrecisionFormula}
\end{equation}

\begin{equation}
\text{Recall} = \frac{\text{True Positive}}{\text{True Positive} + \text{True Negatives}}
\label{equ:RecallFormula}
\end{equation}

\begin{equation}
\text{F1 Score} = \frac{\text{2} * \text{(Precision} * \text{Recall)}}{\text{(Precision} * \text{Recall)}}
\label{equ:F1ScoreFormula}
\end{equation}

\begin{center}
\begin{align*}
    \text{Precision}_1 = \frac{{19}}{19+3} = 0.8636
    \\
    \text{Recall}_1 = \frac{{19}}{19+4} = 0.8261
    \\
    \text{F1 Score}_1 = \frac{{2 * (0.8636 * 0.8261)}}{(0.8636 + 0.8261)} = 0.8444
\end{align*}
\end{center}

\begin{center}
\begin{align*}
    \text{Precision}_2 = \frac{{13}}{13+1} = 0.9286
    \\
    \text{Recall}_2 = \frac{{13}}{13+8} = 0.6190
    \\
    \text{F1 Score}_2 = \frac{{2 * (0.9286 * 0.6190)}}{(0.9286 + 0.6190)} = 0.7428
\end{align*}
\end{center}

\begin{center}
\begin{align*}
    \text{Precision}_3 = \frac{{6}}{6+1} = 0.8571
    \\
    \text{Recall}_3 = \frac{{6}}{6+4} = 0.6000
    \\
    \text{F1 Score}_3 = \frac{{2 * (0.8571 * 0.6000)}}{(0.8571 + 0.6000)} = 0.7059
\end{align*}
\end{center}

\begin{center}
\begin{align*}
    \text{Precision}_4 = \frac{{7}}{7+0} = 1.0000
    \\
    \text{Recall}_4 = \frac{{7}}{7+0} = 1.0000
    \\
    \text{F1 Score}_4 = \frac{{2 * (1.0000 * 1.000)}}{(1.0000 + 1.0000)} = 1.0000
\end{align*}
\end{center}

\begin{center}
\begin{align*}
    \text{Precision}_5= \frac{{13}}{13+2} = 0.8667
    \\
    \text{Recall}_5 = \frac{{13}}{13+2} = 0.8667
    \\
    \text{F1 Score}_5 = \frac{{2 * (0.8667 * 0.8667)}}{(0.8667 + 0.8667)} = 0.8667
\end{align*}
\end{center}

\begin{align*}
\text{Avg Prec} &= \frac{\text{Prec}_1 + \text{Prec}_2 + \text{Prec}_3 + \text{Prec}_4 + \text{Prec}_5}{5} \\
\\
&= \frac{0.863 + 0.928 +  0.857 + 1.000 +  0.866}{5} \\
&= 0.9032
\end{align*}

\begin{align*}
\text{Avg Rec} &=\frac{\text{Rec}_1 + \text{Rec}_2 + \text{Rec}_3 + \text{Rec}_4 + \text{Rec}_5}{5}
\\
&= \frac{0.826 + 0.619 + 0.600 + 1.000 +  0.866}{5}\\
&= 0.7823\
\end{align*}

\begin{align*}
\text{Avg F1 Score} &= \frac{\text{F1}_1 + \text{F1}_2 + \text{F1}_3 + \text{F1}_4 + \text{F1}_5}{5}
\\
&= \frac{0.844 + 0.7428 + 0.705 + 1.000 +  0.866}{5}\\
&= 0.8319
\label{equ:AverageF1Score}
\end{align*}

\subsection{Website Data Visualization}
Our monitoring system uses Cron, a time-based job scheduler, to schedule a script for recording and processing videos every 5 minutes and 30 seconds. The additional 30 seconds are to give Gstreamer (our recording application) time free the camera to start the next process. However, the scheduled hours for running the monitoring system are limited to sunrise (7 am) and sunset (8 pm). This constraint is imposed because the camera system utilizes, Raspberry Pi V2.1, which lacks night vision capabilities. Therefore, the system is scheduled to operate only during daylight hours when sufficient visibility is available. 

The graphs below show 4 out of the 10 available on the IntelliBeeHive web application to show different time periods and demonstrate the changes ins activity, humidity, CPU temperature, and hive temperature throughout the days/weeks/months.
\\

\begin{figure}[H]
  \begin{minipage}[b]{0.48\textwidth}
    \includegraphics[scale=0.53]{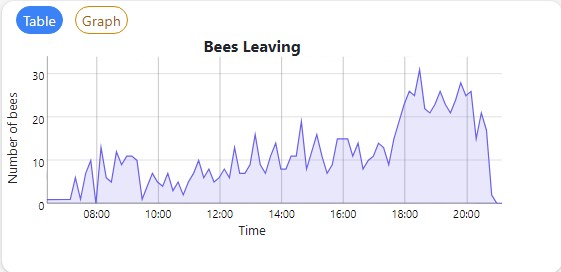}
    \caption[Leaving Graph]{Website graph data for honey bees leaving the hive.}%
    \label{fig:BeeLeavingGraph}
  \end{minipage}
  \hspace{.5cm}
  \begin{minipage}[b]{0.48\textwidth}
    \includegraphics[scale=0.53]{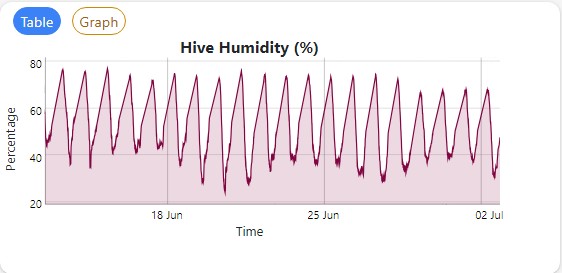}
    \caption[Hive Humidity Graph]{Website graph data for beehive humidity.}%
    \label{fig:HiveHumidityGraph}
  \end{minipage}
  \end{figure}

\begin{figure}[H]
  \begin{minipage}[b]{0.48\textwidth}
    \includegraphics[scale=0.53]{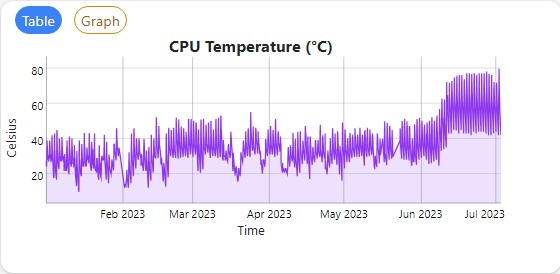}
    \caption[CPU Temp Graph]{Website graph data for CPU Temperature.}%
    \label{fig:CPUTempGraph}
  \end{minipage}
  \hspace{.5cm}
  \begin{minipage}[b]{0.48\textwidth}
    \includegraphics[scale=0.53]{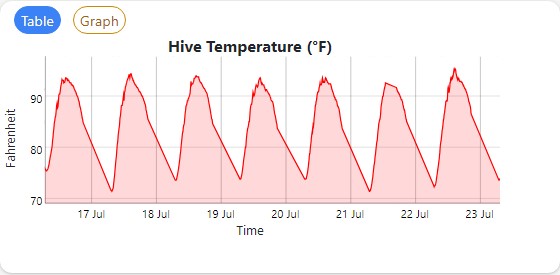}
    \caption[Hive Temp Graph]{Website graph data for beehive temperature.}%
    \label{fig:HiveTempGraph}
  \end{minipage}
\end{figure}

\section{Discussion}\label{ch:Discussion}

In this study, the IntelliBeeHive monitoring system has been successfully designed and implemented using cost-effective technology to ensure accessibility for apiaries of different scales, including hobbyists, commercial businesses, and researchers. Using this monitoring system, users can effectively and efficiently monitor the well-being and behavioral patterns of honey bee hives by analyzing the honey bees' activity with our YOLOv7-tiny models and tracking algorithm. The performance of our IntelliBeeHive system has demonstrated its effectiveness in monitoring the honey bee's activity, achieving an accuracy of 96.28\% in tracking and our pollen model achieved an F1 score of 0.8319.

Future work can be done to further improve and expand our monitoring system. One significant implementation that should be added is the inclusion of real mite data to make our monitoring system fully functional. Additionally, a potential improvement could be upgrading our camera to support night vision. While night vision is not currently necessary since honey bees are inactive at night, a night vision-capable camera would enable our monitoring system to run continuously. 

Another major step for the future would be to collaborate with beekeepers and deploy our monitoring system in beehives from various locations around the world. This testing will help evaluate the system's overall performance in diverse and unpredictable environments, such as dealing with challenges like extreme heat when deployed in Texas. Using the feedback from other beekeepers we can fine-tune our design to make our monitoring system more robust and reliable.

\section*{Acknowledgements}
This work was supported by the USDA National Institute of Food and Agriculture (Award No. 2019-68017-29694).
We'd like to extend special thanks to our webmaster Juan Velazquez, who helped manage our website.

\ifCLASSOPTIONcaptionsoff
  \newpage
\fi

\bibliographystyle{IEEEtran}

\bibliography{bibtex/bib/refcite}

\end{document}